\documentclass[oneside,10pt]{article}
\usepackage[a4paper,width=150mm,top=25mm,bottom=25mm]{geometry}
\usepackage[utf8]{inputenc}
\usepackage{authblk}
\usepackage{hyperref}
\usepackage{graphicx}
\usepackage{xspace}
\usepackage[frozencache,cachedir=.]{minted}
\usepackage[dvipsnames,table]{xcolor}
\usepackage[square,numbers,sort&compress]{natbib}
\usepackage{array}

\newcommand{\incode}[1]{\texttt{#1}}

\newcommand{\ie}{\emph{i.e.},\xspace}
\newcommand{\eg}{\emph{e.g.},\xspace}
\newcommand{\cf}{\emph{cf.}\xspace}
\newcommand{\zennit}{\emph{Zennit}\xspace}
\newcommand{\corelay}{\emph{CoRelAy}\xspace}
\newcommand{\virelay}{\emph{ViRelAy}\xspace}

\newcommand{\composite}{\texttt{Composite}\xspace}
\newcommand{\composites}{\texttt{Composites}\xspace}

\newcommand{\canonizers}{\texttt{Canonizers}\xspace}

\usepackage{glossaries}
\newacronym{dtd}{DTD}{Deep Taylor Decomposition}
\newacronym{dnn}{DNN}{Deep Neural Network}
\newacronym{lrp}{LRP}{Layer-wise Relevance Propagation}
\newacronym{spray}{SpRAy}{Spectral Relevance Analysis}
\newacronym{tsne}{t-SNE}{t-distributed Stochastic Neighborhood Embedding}
\newacronym{xai}{XAI}{Explainable Artificial Intelligence}

\title{Software for Dataset-wide XAI: From Local Explanations to Global Insights with Zennit, CoRelAy, and ViRelAy}
\author[1,2]{Christopher J. Anders}
\author[3]{David Neumann}
\author[2,3,4]{Wojciech Samek}
\author[1,2,5,6]{Klaus-Robert M\"uller}
\author[3]{Sebastian Lapuschkin}
\affil[1]{Machine Learning Group, Department of Electrical Engineering and Computer Science, Technische Universit\"at Berlin, Germany}
\affil[2]{BIFOLD -- Berlin Institute for the Foundations of Learning and Data, Berlin, Germany}
\affil[3]{Department of Artificial Intelligence, Fraunhofer Heinrich Hertz Institute, Berlin, Germany}
\affil[4]{Machine Learning and Communications Group, Department of Electrical Engineering and Computer Science, Technische Universit\"at Berlin, Germany}
\affil[5]{Department of Artificial Intelligence, Korea University, Seoul, Republic of Korea}
\affil[6]{Max Planck Institut f\"ur Informatik, Saarbr\"ucken, Germany}

\date{}

\begin{document}

\maketitle

\begin{abstract}
    Deep Neural Networks (DNNs) are known to be strong predictors, but their prediction strategies can rarely be understood.
    With recent advances in Explainable Artificial Intelligence (XAI), approaches are available to explore the reasoning behind those complex models' predictions.
    Among post-hoc attribution methods, Layer-wise Relevance Propagation (LRP) shows high performance.
    For deeper quantitative analysis, manual approaches exist, but without the right tools they are unnecessarily labor intensive.
    In this software paper, we introduce three software packages targeted at scientists to explore model reasoning using attribution approaches and beyond:
    (1) \emph{Zennit} -- a highly customizable and intuitive attribution framework implementing LRP and related approaches in PyTorch,
    (2) \emph{CoRelAy} -- a framework to easily and quickly construct quantitative analysis pipelines for dataset-wide analyses of explanations,
    and
    (3)~\emph{ViRelAy} -- a web-application to interactively explore data, attributions, and analysis results.
    With this, we provide a standardized implementation solution for XAI, to contribute towards more reproducibility in our field.
\end{abstract}

\section{Introduction}
    There is no doubt that \glspl{dnn} are strong predictors,
    which are able to solve many problems~(\eg \citep{senior2020improved, unke2021spookynet, dosovitskiy2021image}).
    With their inherent complexity, however, also comes a heavy down-side, which is the lack of transparency of \glspl{dnn}.
    Recent advances in \gls{xai} (see, \eg \citep{guidotti2019survey,samek2019explainable,arrieta2020explainable,samek2021toward,holzinger2022explainable} for a timely overview), however, allow for a more in-depth investigation of \gls{dnn} behavior.
    Here, attribution methods are able to yield local explanations, \ie attribution scores for all (input) features of individual samples.
    The \gls{lrp}~\citep{bach2015pixel,montavon2019layer},
    for example,
    with its mathematical roots in \gls{dtd}~\citep{montavon2017explaining} and its various purposed modified backpropagation rules~\citep{montavon2019layer, kohlbrenner2020towards, arras2019explaining},
    has proven to be a particularly powerful method of local \gls{xai} showing excellent results~\citep{samek2017evaluating, poerner2018evaluating, kohlbrenner2020towards, arras2020towards} when recommended guidelines are followed,
    yet it is rarely used \emph{to its full potential}, \eg due to a lack of ready-made and \emph{complete} (see Table \ref{tab:attribution_frameworks}) implementations.
    In particular, an exhaustive implementation of \gls{lrp} following contemporary recommendations from literature~\citep{montavon2019layer,kohlbrenner2020towards,samek2021toward} is still lacking for the popular PyTorch framework.
    As one of our contributions, we thus aim to make a \emph{versatile and flexible} implementation of \gls{lrp} available to the community,
    which goes beyond the simple variants \gls{lrp}-$\varepsilon$ or (Gradient$\times$Input) often provided as the sole~\citep{kokhlikyan2020captum}, yet not universally recommended variants, of the method.

    If employed correctly, local \gls{xai}
    has the potential to point out previously unknown but interesting model behavior, or biased and artifactual predictions~\citep{lapuschkin2016analyzing, aeles2021revealing}.
    With very large datasets however, a thorough (manual) analysis of attribution results, \eg for the understanding and verification of model behavior, or the discovery of systematic misbehavior are very labor- and time-intensive.
    Still, further insight beyond local attributions is required, \eg to understand global model behavior, or to notice systematic Clever Hans~\citep{pfungst1911clever,lapuschkin2019unmasking} traits of a model.
    Recent approaches such as \gls{spray}~\citep{lapuschkin2019unmasking,anders2022finding} provide a solution to this arduous task by automating large parts of the analysis workload and are thus, together with appropriate visualizations, aiding in the discovery of prediction strategies employed by a \gls{dnn} model.

    In this paper,
    we introduce three software packages targeted at scientists to explore the reasoning of machine learning models based on dataset-wide \gls{xai}:
    \begin{enumerate}
        \item With \zennit we provide a highly customizable, yet intuitive local \gls{xai} framework, for PyTorch. It is focused on rule-based approaches such as \gls{lrp} and based on PyTorch's \incode{Module} structure, enabling (and delivering) implementations of various attribution methods.
        \item \corelay in turn digests attributions (and possibly also other sources of data), and can be used to quickly build elaborate, dataset-wide analysis pipelines such as \gls{spray}, consisting of, \eg processing, clustering and embedding steps.
        The framework aims at efficiency during analysis by re-using matching (partial) pipeline results as often as possible within and between pipeline executions, instead of re-computing the complete pipeline each time, \eg due to parameter changes.
        \item \virelay provides a user-friendly entry point to the analysis results from \zennit and \corelay in form of an interactive web-application. During the exploration of data with model attributions, clusterings, and visualizable embeddings, researchers can import, export, bookmark, and share particular findings with their peers.
    \end{enumerate}

    In combination, these three tools enable \gls{xai} to be used to quantitatively and qualitatively explore and investigate large scale models and data:
    Local model explanations can be obtained through attributions computed with
    \zennit\footnote{\url{https://github.com/chr5tphr/zennit}, documentation at \url{https://zennit.rtfd.io}}.
    Users may then analyze large sets of attributions computed over whole datasets with pipelines built in
    \corelay\footnote{\url{https://github.com/virelay/corelay}, documentation at \url{https://corelay.rtfd.io}},
    of which the results can then be visualized and investigated with
    \virelay\footnote{\url{https://github.com/virelay/virelay}, documentation at \url{https://virelay.rtfd.io}}.
    The insights obtainable through this particular, yet flexible recipe allows to go beyond passively observant \gls{xai},
    \eg by fuelling a strategy of informed intervention; only through the use of the here introduced scalable software packages, we were able to identify systematically biased reasoning in \gls{dnn} models trained on ImageNet~\citep{anders2022finding}.

    All three of these tools are thoroughly tested, documented, provide continuous integration and encourage contributions.
    Since its inception, \zennit has considerably grown, with a community increasing in size with multiple contributors, new releases, pull requests and use in other software.

\paragraph{Related Work}
Multiple software frameworks have been introduced using different deep learning libraries to compute model attributions.
One of the earlier and comprehensive \gls{xai} software packages is the LRP Toolbox~\cite{lapuschkin2016toolbox},
providing implementations of a wide array of recommended \gls{lrp} decomposition rules
for the Caffe Deep Learning Framework~\citep{jia2014caffe},
as well as Matlab and Python (using NumPy~\citep{harris2020array} and CuPy~\citep{okuta2017cupy}) via custom neural network interfaces.
The software framework iNNvestigate~\citep{alber2019innvestigate}, which is based on TensorFlow~\citep{abadi2016tensorflow} and Keras~\citep{chollet2015keras}, implements LRP and other attribution approaches.
While it provides a straight-forward approach to apply multiple attribution methods on existing Keras models, its structure makes customization (\eg by implementing custom rules and compositions of rules) non-trivial.
Captum~\citep{kokhlikyan2020captum}, which is tightly integrated into PyTorch, provides a broad spectrum of attribution methods. It is very customizable, but lacks specificity for layer-type specific implementations of decomposition rules necessary for \gls{lrp}, thus requiring a lot of work to use state-of-the-art defaults for \gls{lrp}.
TorchRay~\citep{fong2019understanding} is another attribution framework built on PyTorch, which also provides a broad spectrum of attribution methods, but has no support for LRP.
OpenXAI~\citep{agarwal2022openxai} and Quantus~\citep{hedstrom2023quantus} provide tools to evaluate XAI methods, striving for reproducibility in XAI research in line with the goal of our proposed packages.
\zennit has proven its flexibility as the foundation of the design of novel XAI approaches in the implementation of concept relevance propagation \citep{achtibat2022from}.

\section{Attribution with Zennit}
\label{section:attribution-with-zennit}
\zennit provides a framework for attribution in PyTorch~\citep{paszke2019pytorch}.
It is based on the \incode{Module} structure in PyTorch, and makes heavy use of its \incode{Autograd} and \incode{Hook} functionalities.
It is mainly focused on implementing the rule-based approach used by \gls{lrp}~\citep{bach2015pixel} in a simple and intuitive manner:
The provision of an easy to modify and flexible implementation of \gls{lrp} is paramount for obtaining excellent results,
by optimally aligning the method to the characteristics of the model (or parts thereof) to be analyzed~\cite{kohlbrenner2020towards,montavon2019layer,samek2021toward}.

Simpler attribution methods, such as SmoothGrad~\citep{smilkov2017smoothgrad} and Integrated Gradients~\citep{sundararajan2017axiomatic}, are also implemented, although they do not make use of the rule-based system, but are straight forward functions of the gradient of the model to be analyzed.

\paragraph{Rule-Based Attributions}
Rule-based attribution methods assign different rules to \incode{Modules} within a model depending on the function and context.
In \zennit, rule-based attributions are computed by attaching \emph{forward} and \emph{backward} \incode{Hooks} to \incode{Modules} (layers), such that computing the gradient of the model will instead provide the desired attribution.
At the heart of \zennit is the \incode{BasicHook}, which contains the functionality to register and remove modifications to a single \incode{Module} (layer), and a general attribution method.
\emph{Rules} are created by providing functions to a \incode{BasicHook} to customize the general attribution method with modified inputs, parameters, and accumulators.
This makes an implementation of new rules trivial.
All popular rules for LRP (for an overview see~\citep{montavon2019layer}), as well as others, such as GuidedBackprop~\citep{springenberg2015striving} and ExcitationBackprop~\citep{zhang2018neural}, come pre-implemented.

\paragraph{Mapping Rules with Composites}
The biggest challenge when aiming at a successful implementation of rule-based attribution methods is to assign the desired rules to all individual layers.
\zennit solves this by implementing \composites, which are mappings from \incode{Module}-properties to rules.
\incode{Module}-properties are for example the name or type of function, its (hyper-)parameters or its position within the model.
\composites are provided with a \incode{module\_map}, which, given the \incode{Module}-properties, returns a template-rule to be assigned to the layer.
One example for a built-in basic \composite is the \incode{SpecialFirstLayerMapComposite}, which assigns rules based on layer types, but handles the first linear layer differently.
This is the basis for most \gls{lrp}-based \composites for feed-forward networks, like \incode{EpsilonGammaBox}, which uses the \gls{lrp}-$\varepsilon$-rule for dense layers, the \gls{lrp}-$\gamma$-rule for convolutional layers, and the \gls{lrp}-$Z^B$-rule (or box-rule) for the first convolutional layer~\citep{montavon2019layer}.

\paragraph{Temporary Model Modification with Canonizers}
Another challenge with rule-based attribution methods is that their rules may not directly be applicable to many networks, unless they are transformed into a canonical form~\citep{yeom2021pruning, ruff2021unifying, samek2021toward}.
For example, multiple consecutive linear layers with only one activation at the very end cannot always be trivially attributed with all variants of \gls{lrp} unless the consecutive linear layers are merged into a single one.
Recent work has also shown that appropriate network canonization can have a perceivably and measurably positive impact on the quality of the attribution \citep{motzkus2022measurably,pahde2022optimizing}.
A common example for this structure is batch normalization~\citep{ioffe2015batch}.
To temporarily modify models in-place into a canonical form, \zennit implements \incode{Canonizers}.
Due to its common application, \zennit provides the \incode{MergeBatchNorm} \incode{Canonizer}, to temporarily merge batch normalization layers into an adjacent linear layer~\citep{hui2019batchnorm,alber2019efficient,guillemot2020breaking}.
A general \incode{Canonizer}, which is, for example, needed to apply \gls{lrp} on ResNet~\citep{he2016deep,zagoruyko2016wide}, is the \incode{AttributeCanonizer}, which, while registered, will modify (instance) attributes in place, for example, to split a single module for which there is no rule, into multiple ones for which rules may then be assigned.
Model-specific \incode{Canonizers} for popular models like VGG-16~\citep{simonyan2014very} and ResNet~\citep{he2016deep,zagoruyko2016wide} from, \eg Torchvision~\citep{marcel2010torchvision}, are implemented for convenience.
\incode{Canonizers} are directly provided to \composites, so they will be applied right before the rules are mapped to the layers when registering the \composite to a model.

\paragraph{Attributors}
Attributors provide optional convenience functionality to either compute the gradient given a model and a \composite, or to implement black-box attribution approaches such as SmoothGrad and Integrated Gradients.
Given a gradient-based black-box attribution approach, \eg SmoothGrad, it is also possible to supply a \composite, to compute a combination of, \eg \gls{lrp} and SmoothGrad, since the composite will modify the gradient of the model.
Non-gradient based approaches, like Occlusion Analysis~\citep{zeiler2014visualizing}, cannot be combined with \composites, since the modified gradient of the \composite has no effect on the result.

\begin{figure}[t]
    \centering
    \includegraphics[width=\textwidth]{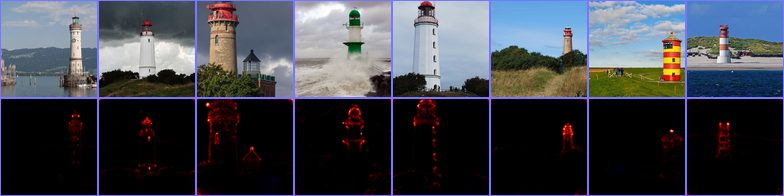}
    \caption{
        Heatmaps of attributions of lighthouses, using the pre-trained VGG-16 network provided by Torchvision.
        The \composite \incode{EpsilonGammaBox} was used and the attributions were visualized with the color map \incode{coldnhot} (negative relevance is light-/blue, irrelevant pixels are black, positive relevance is red to yellow).
    }
    \label{fig:lighthouses}
\end{figure}

\paragraph{Heatmaps}
Since attributions for image data are often visualized in heatmaps, \zennit comes with an image module to easily visualize and store attributions as heatmap images.
Various color maps are available.
The images are stored using intensities and 8-bit palettes where indices correspond to the attributed relevances.
This makes it easy to change the color map afterwards, without re-computing the relevance values.
An example for visualized heatmaps is given in Figure \ref{fig:lighthouses}.

\hfill

Listing \ref{lst:zennit} shows a typical application of \zennit on the Torchvision VGG16 model with BatchNorm.
The \gls{lrp} attribution is implemented by passing the \incode{low} and \incode{high} keyword arguments (i.e. the lowest and highest possible input value)
as well as the neccessary \canonizers (here \incode{SequentialMergeBatchNorm}) to the built-in \incode{EpsilonGammaBox} \composite,
and then passing the composite as well as the model to the \incode{Gradient} \texttt{Attributor}.

\begin{listing}[ht]
    \centering
    \begin{minted}[
        fontsize=\footnotesize,
        frame=single
    ]{python}
        import torch
        from torchvision.models import vgg16_bn

        from zennit.composites import EpsilonGammaBox
        from zennit.canonizers import SequentialMergeBatchNorm
        from zennit.attribution import Gradient

        # create some demonstrative random data
        data = torch.randn(1, 3, 224, 224)
        # create a randomly initialized VGG16 model with BatchNorm
        model = vgg16_bn()

        # list of canonizers; merges BatchNorms into adjacent linear layers
        canonizers = [SequentialMergeBatchNorm()]
        # create the composite; pass parameters for LRP rules and canonizers
        composite = EpsilonGammaBox(low=-3., high=3., canonizers=canonizers)

        # enter context, applying canonizers and registering all rules
        with Gradient(model=model, composite=composite) as attributor:
            # define relevance at output, i.e. attribution wrt. class
            relevance_at_output = torch.eye(1000)[[0]]
            # pass input and output relevance to attributor
            # computes the gradient modified by the composite
            output, relevance = attributor(data, relevance_at_output)
    \end{minted}
    \caption{
        Example Python code to compute \gls{lrp} heatmaps of random data of Torchvision's VGG16 model with BatchNorm.
        The \incode{SequentialBatchNorm} \incode{Canonizer} merges the BatchNorm into adjacent linear layers, before the \incode{EpsilonGammaBox} \incode{Composite} is applied.
        The \incode{Gradient} \incode{Attributor} computes the gradient, which is modified by the \incode{composite} within its context, resulting in the computation of the \gls{lrp} attribution with the best-practice \incode{EpsilonGammaBox} rule-set.
    }
    \label{lst:zennit}
\end{listing}

For more code examples, how-tos and an in-depth tutorial on \zennit, we refer to the documentation\footnote{\url{https://zennit.rtfd.io/en/0.5.1/getting-started.html}}.

\section{Building Analysis Pipelines with CoRelAy}
\label{section:building-analysis-pipelines-with-corelay}
While attribution methods can give a qualitative insight into a model's prediction strategies,
a user may only guess how the attributions of individual heatmaps are part of the model's reasoning.
A deeper insight into the model may be gained by conducting a dataset-wide analysis.
\citet{lapuschkin2019unmasking} introduced \glsdesc{spray}, with which they quantitatively analyze a model's prediction strategy by visually embedding and clustering attributions with Spectral Clustering~\citep{meila2001random, ng2002spectral} and \gls{tsne}~\citep{maaten2008visualizing}.
\citet{anders2022finding} extended \gls{spray} by using different clustering and visual embeddings, as well as computing a pre-ranking of interesting classes based on the linear separability of their clusterings.
\corelay is a tool to quickly compose quantitative analysis pipelines like \gls{spray}, which provide multiple embeddings, representations, and labels of the data.
While our main use-case and motivation for \corelay was to analyze attributions provided by \zennit, \corelay is not limited to any kind of data, \eg \corelay may also be used for a quick dataset exploration with multiple clusterings and embeddings.

\begin{listing}[ht]
    \centering
    \begin{minted}[
        fontsize=\footnotesize,
        frame=single
    ]{python}
    import h5py
    import numpy as np

    from corelay.processor.flow import Sequential, Parallel
    from corelay.pipeline.spectral import SpectralClustering
    from corelay.processor.clustering import KMeans
    from corelay.processor.embedding import TSNEEmbedding, EigenDecomposition
    from corelay.io.storage import HashedHDF5

    # open HDF5 file in append mode
    with h5py.File('spray.h5', 'a') as fd:
       # io-object to store outputs based on hash in HDF5
       iobj = HashedHDF5(fd.require_group('proc_data'))
       # example for a pre-defined pipeline
       pipeline = SpectralClustering(
           # processors can be assigned to pre-defined tasks
           embedding=EigenDecomposition(n_eigval=8, io=iobj),
           # combine multiple processors with Parallel; copy the input
           # as many times as there are Processors with broadcast=True
           clustering=Parallel([
               Parallel([
                   # multiple k-means clusterings with different k
                   KMeans(n_clusters=k, io=iobj) for k in range(2, 20)
               ], broadcast=True),
               # pass an io-object to store intermediate results
               TSNEEmbedding(io=iobj)
               # Processors with is_output=True will be outputs
           ], broadcast=True, is_output=True)
       )
       data = numpy.random.normal(size=(64, 3, 32, 32))
       # execute the pipeline, producing all outputs
       clusterings, tsne = pipeline(data)
    \end{minted}
    \caption{Example code to instantiate and execute a simple \gls{spray} pipeline, using $8$ eigenvalues for the Spectral Embedding, clustering using k-means with $k\in\{2,...,20\}$, and visualizing with \gls{tsne}. The results are additionally cached in a file called \incode{spray.h5}.}
    \label{lst:pipeline}
\end{listing}

\paragraph{Processors and Params}
\incode{Processors} are the actions in a pipeline.
To implement a \incode{Processor}, an inheriting class will have to implement a method with the name \incode{function}, and class-scope \incode{Params}.
In Python terminology, \incode{Params} are descriptors, which change based on the instance they are bound to (similar to methods).
\incode{Params} are used to easily define the arguments of \incode{Processors}, their desired types, default values, and others.
\incode{Processors} already have the \incode{Params} \incode{is\_output}, to signal that the output of this \incode{Processor} should be returned by the \incode{Pipeline} (even if intermediate), and \incode{io}, which can be assigned to a \incode{Storage} object to cache data on disk.
Many \incode{Processors} come pre-implemented with \corelay, which are categorized into pre-processing, distance functions, affinity functions, Laplacians, embedding methods, and flow \incode{Processors}.
Flow \incode{Processors} are used to design more complex flows of \incode{Pipelines}, of which the most important are \incode{Parallel} and \incode{Sequential}.
With \incode{Parallel}, the output of the previous \incode{Processor} may be passed to multiple \incode{Processors}, \eg to compute multiple clusterings on the same data or to try to compute a visual embedding with different hyperparameters.
With \incode{Sequential}, \incode{Processors} may be combined to do multiple steps where there is only a single \incode{Task} in a \incode{Pipeline}.

\paragraph{Pipelines and Tasks}
\incode{Pipelines} are feed-forward functions, which have \incode{Tasks} that have to be fulfilled from front to back to execute the pipeline.
In \corelay, \incode{Pipelines} can be seen as computation templates, where there are steps involved to compute a certain result, which can be individually changed.
A \incode{Task} is such a step, with a default \incode{Processor}, and optionally an \emph{allowed type} of \incode{Processor}.
When instantiating a \incode{Pipeline}, \incode{Tasks} may be assigned a new \incode{Processor} to handle the data instead of the default one.
A \incode{Pipeline} can be executed by simply calling it as a function with the input data as its arguments.
Depending on the \incode{Processors} used and their respective \incode{is\_output} flags, the output of the \incode{Pipeline} may have none, one, or a hierarchy of results.
If \incode{Processors} within the \incode{Pipeline} own an \incode{io} object, they will cache their results by hashing the input data and parameters.
When calling the same \incode{Pipeline} with the same data, these results will be looked up instead of being re-computed.
\corelay has a \gls{spray} \incode{Pipeline} (\cf \citep{anders2022finding}) pre-implemented, to produce data which can be directly used with \virelay.
A \gls{spray} \incode{Pipeline} may be instantiated and executed as shown in Listing \ref{lst:pipeline}.

\section{Interactive Visualization with ViRelAy}
With quantitative analyses, a large amount of results are created, and it may become hard to connect the different results and representations with the original data.
A labor-intensive manual comparison and creation of individual plots, in an attempt to extract the essence of the results may become inevitable to find correlations in the data.
The analysis performed with \gls{spray} has a very distinct and common set of objects that need to be compared:
the source data points,
their attributions (wrt.~a model),
a visual 2-dimensional representation of the (embedded) attribution data,
clustering labels and global auxiliary scores.
\virelay is an interactive web-application, with which the results may be freely explored by visually connecting these 5 objects.
\virelay's back-end is implemented in Python using Flask~\citep{grinberg2014flask}, and its front-end is implemented using Angular~\citep{jain2014angularjs}.

\paragraph{Data Loading}
\virelay is designed to process the data of \corelay.
The results of \corelay are stored in HDF5~\citep{fortner1998hdf} files in a hierarchy that \virelay is able to use post-hoc, reducing loading times for an improved user interaction quality.
The analysis file, along with the source data and the attribution data, both also stored in HDF5, are referenced in a project file.
A single project file may contain one source dataset with one attribution for each sample, as well as an arbitrary amount of analysis files.
To compare different datasets or attribution approaches, \virelay can be executed by supplying an arbitrary amount of project files, between which the client may switch during execution.

\begin{figure}[ht]
    \centering
    \includegraphics[width=\textwidth]{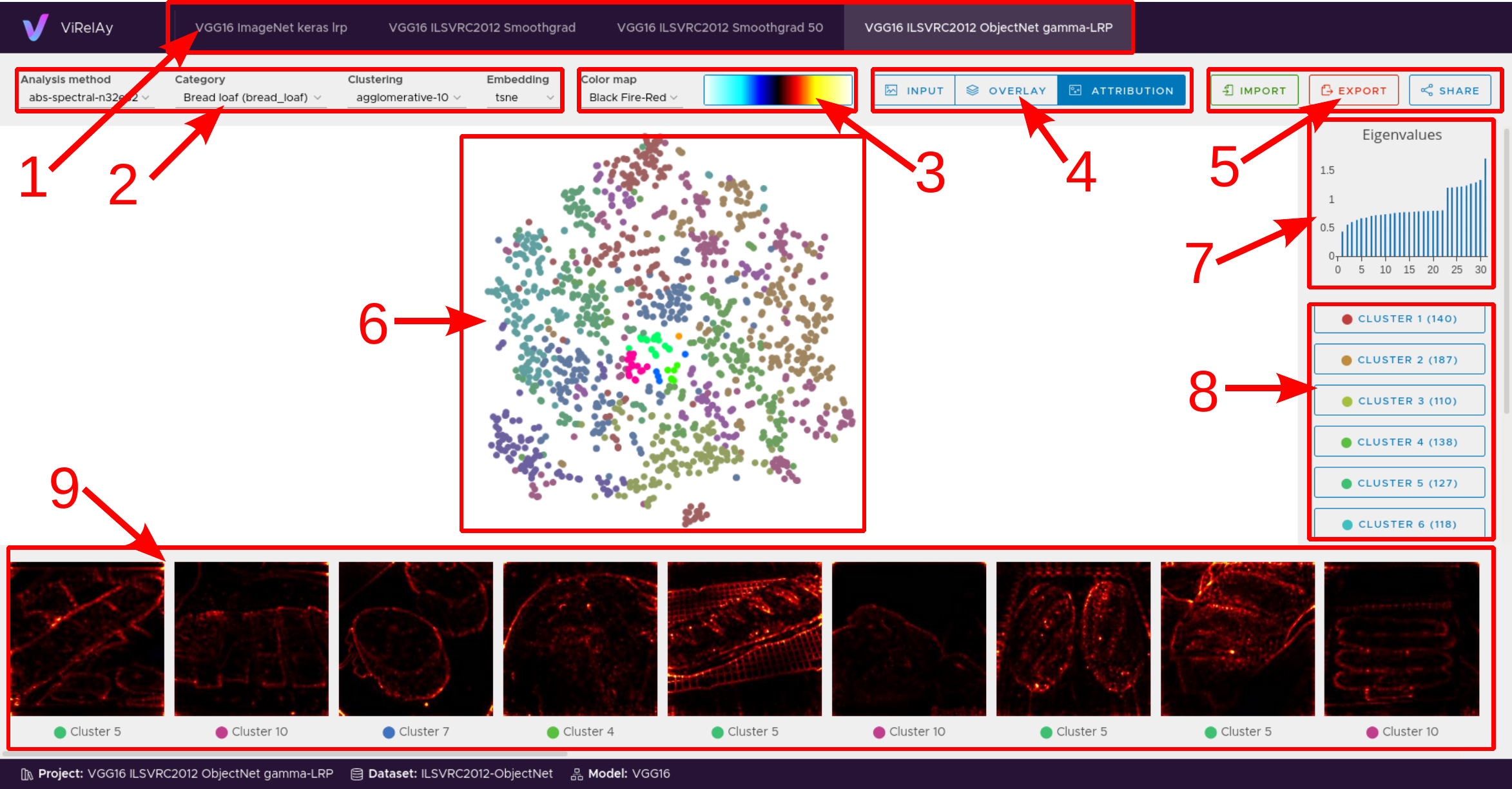}
    \caption{
        The \virelay user interface.
        Highlighted points are:
        (1) Project selection,
        (2) analysis setup and category selection,
        (3) color map selection,
        (4) data/attribution visualization mode selection,
        (5) import/export/share current selection,
        (6) 2d visual embedding canvas,
        (7) auxiliary score plot,
        (8) cluster point selection,
        (9) data/attribution visualization.
    }
    \label{fig:virelay}
\end{figure}

\paragraph{Explorative User Interaction}
The user interface is shown in Figure \ref{fig:virelay}.
At the top of the interface is (1) the project selection, where the projects, as dictated by the project files, show up as tabs and may be selected to switch between datasets and attribution methods.
Below the project selection, on the left side is (2) the analysis selection, where the analysis approach (given by supplying multiple analysis files in a single project file), the category (which often is the data label, but may be chosen as any group of data points), the clustering method (which influences (8) the available clusters and (6) the data point coloring), and the embedding (which is the 2d representation of the data points as shown in (6) the visualization canvas) can be selected.
Selecting a different analysis method resets all categories.
To the right is (3) the color map selection, which changes the color map used in (9) the data/attribution selection, with a color bar indicating low (left) and high (right) values.
The next item to the right is (4) the data/attribution visualization mode selection, which changes whether (9) the data/attribution visualization shows the source data (input), its attribution with the selected color map (attribution), or the attribution superimposed onto a gray-scale image of the source data (overlay).
The (5) \emph{import} and \emph{export} buttons allow to export the currently selected analysis, category, clustering, embedding, color map, visualization mode and selected points by downloading a JSON-file, or importing a JSON-file to change the selections to the configuration of a previously downloaded file.
This may be used either to remember or to share interesting results.
The selection may also be shared or bookmarked in the form of a URL using the (5) \emph{share} button.
At the center of the interface is (6) the 2d-visualization canvas, which shows the points in the selected 2-dimensional embedding space (produced by, \eg \gls{tsne}) colored by the clusters indicated in (8) the cluster point selection.
In this canvas, the user may zoom or pan, and select points which will be highlighted by a more saturated color and shown in (9) the data/attribution visualization.
Hovering over data points will show a preview of the source data inside the canvas.
To the right is (7) the auxiliary category score plot, which in Figure~\ref{fig:virelay} are the eigenvalues of the Spectral Embedding.
Below, there is (8) the cluster point selection, which shows the available clusters of the selected clustering, as well as the colors used for members of these clusters in (6) the 2d-visualization canvas, and the number of points in this cluster in parentheses.
Finally, at the bottom is (9) the data/attribution visualization, where, depending on which mode was selected in (4) the data/attribution mode selection, will show either the source data, the attribution heatmap, or the attribution superimposed on a gray-scale version of the source image, of a subset of the selected points.

\section{Comparison to Alternative Frameworks}
To put our work into context, we provide a comparison of each of our frameworks to available alternatives.
We compare \zennit to similar attribution frameworks, with a focus on propagation based approaches, in particular \gls{lrp}.
Although \corelay is a domain-specific framework, we compare it to a few alternatives with increasing complexity.
Since \virelay solves a very specific visualization problem for which no other frameworks exist, we give an overview of other visualization frameworks for \gls{xai}.

\paragraph{Attribution Frameworks}
\renewcommand{\arraystretch}{1.5}
\definecolor{slightgray}{gray}{0.9}

Accompanying the rise of \gls{xai}, there are many explainability frameworks beside \zennit for various areas of application.
Some (e.g. Captum Insights) even provide visual front ends to enhance interpretability.
Although multiple frameworks seem to solve similar challenges, some projects did not stand the test of time, and ultimately became unmaintained shortly after their publication.
Table \ref{tab:attribution_frameworks} lists some popular attribution frameworks,
along \zennit, with columns that focus on the main objective of \zennit: to provide a feature-complete,
modular and customizable framework for propagation-based attribution methods,
focused on LRP, with additional general attribution method capabilities.
Although many frameworks primarily designed for explanation of classical or white-box (glass-box) methods (interpretML \citep{nori2019interpretml}, explainerdashboard \citep{dijk2022explainerdashboard}, alibi \citep{klaise2021alibi})
have some overlap in their approaches which are also commonly used for \gls{dnn},
we do not list them here due to their limited comparability.

\begin{table}[ht]
    \centering
    \resizebox{\textwidth}{!}{
    \begin{tabular}{|m{7.5em}|m{6em}|m{12em}|m{6.5em}|m{12em}|m{8em}|} \hline \rowcolor{slightgray}
            \textbf{Framework} &
            \textbf{Back-end} &
            \textbf{Propagation\newline Attribution} &
            \textbf{Propagation\newline Rule-map} &
            \textbf{Other Attribution\newline (Notable)} &
            \textbf{Documentation\newline Tests}
        \\ \hline
            Zennit\newline (ours) &
            PyTorch &
                Common \gls{lrp}~\citep{montavon2019layer}\newline
                Uncommon/Custom \gls{lrp}\newline
                Guided Backprop~\citep{springenberg2015striving}\newline
                Excitation Backprop~\citep{zhang2018neural}&
            Built-In\newline Custom\newline Canonization&
                SmoothGrad~\citep{smilkov2017smoothgrad}\newline
                Integrated Gradients~\citep{sundararajan2017axiomatic}\newline
                Occlusion~\citep{zeiler2014visualizing}&
            Full Usage \newline API \newline Tutorials\newline Fully Tested + CI
        \\ \hline
            Captum~\citep{kokhlikyan2020captum} &
            PyTorch &
                \gls{lrp}-$\varepsilon$~\citep{montavon2019layer}\newline
                DeepLIFT(+Shap)~\citep{shrikumar2017learning,lundberg2017unified}\newline
                Guided Backprop~\citep{springenberg2015striving}&
            None &
                SmoothGrad~\citep{smilkov2017smoothgrad}\newline
                Integrated Gradients~\citep{sundararajan2017axiomatic}\newline
                Conductance~\citep{dhamdhere2019important,shrikumar2018computationally}\newline
                GradientShap~\citep{lundberg2017unified}\newline
                KernelShap~\citep{lundberg2017unified}\newline
                GradCAM~\citep{selvaraju2020grad}\newline
                Occlusion~\citep{zeiler2014visualizing}\newline
                LIME~\citep{ribeiro2016should}\newline
                Shapley Values \citep{castro2009polynomial,strumbelj2010efficient}&
            Full Usage \newline API \newline Tutorials\newline Fully Tested + CI
        \\ \hline
            TorchRay~\citep{fong2019understanding}\newline (unmaintained) &
            PyTorch &
                Guided Backprop~\citep{springenberg2015striving}\newline
                Excitation Backprop~\citep{zhang2018neural}&
            None &
                GradCAM~\citep{selvaraju2020grad}\newline
                Occlusion~\citep{zeiler2014visualizing}\newline
                LIME~\citep{ribeiro2016should}\newline
                RISE~\citep{petsiuk2018rise}\newline
                Extremal Perturbation~\citep{fong2019understanding}&
            Joint Usage+API\newline Examples\newline Benchmarks
         \\ \hline
            iNNvestigate~\citep{alber2019innvestigate} &
            Tensorflow/\newline Keras &
                Common \gls{lrp}~\citep{montavon2019layer}\newline
                PatternAttribution~\citep{kindermans2018learning}\newline
                DeepLIFT~\citep{shrikumar2017learning}\newline
                Guided Backprop~\citep{springenberg2015striving}&
            Built-In &
                SmoothGrad~\citep{smilkov2017smoothgrad}\newline
                Integrated Gradients~\citep{sundararajan2017axiomatic}&
            Usage in Readme\newline API\newline Tutorials\newline Fully Tested + CI
        \\ \hline
            DeepExplain\citep{ancona2018towards}\newline (unmaintained) &
            Tensorflow/\newline Keras &
                \gls{lrp}-$\varepsilon$~\citep{montavon2019layer}\newline
                DeepLIFT~\citep{shrikumar2017learning}&
            None &
                Integrated Gradients~\citep{sundararajan2017axiomatic}\newline
                Occlusion~\citep{zeiler2014visualizing}\newline
                Shapley Values \citep{castro2009polynomial,strumbelj2010efficient}&
            Usage in Readme\newline Examples\newline Tests + CI
        \\ \hline
    \end{tabular}
    }
    \caption{
        Python frameworks supporting post-hoc attribution for \gls{xai} of \glspl{dnn}.
        \textbf{Framework} lists the name of the framework.
        \textbf{Back-end} shows the Python-library the framework is based on.
        \textbf{Propagation Attribution} are the supported propagation-based attribution methods of the framework.
        \textbf{Propagation Rule-map} describes the framework's support for mapping different rules to layers or parts of a model.
        \textbf{Other Attribution (Notable)} are the \emph{notable} (i.e. non-trivial), non-propagation-based attribution approaches supported by the framework.
        \textbf{Documentation/Tests} highlights the framework's state of the documentation and tests (with continuous integration).
    }
    \label{tab:attribution_frameworks}
\end{table}

The framework that is most comparable to \zennit is iNNvestigate~\cite{alber2019innvestigate}, which implements common and best-practice rules for \gls{lrp} in TensorFlow~\citep{abadi2016tensorflow} and Keras~\citep{chollet2015keras}.
It is well suited for \gls{lrp} on models implemented in Keras,
but the lack of an easily configurable interface to implement custom rules or rule-maps makes its application on more novel models less efficient.
Although iNNvestigate is the only other framework with any rule-mapping capabilities,
\zennit is the only framework that provides \emph{Canonization} to adapt models where \gls{lrp}-rules would otherwise not be applicable.
Tests and examples with continuous integration as well as a basic usage in the read-me and an API reference are provided with iNNvestigate.

Captum~\cite{kokhlikyan2020captum} implements many common attribution approaches in PyTorch, which makes it comparable to \zennit.
While it provides a wide collection of methods, it only supports simple propagation-based attribution, which relates to its limited implementation of \gls{lrp}.
In Captum, currently only \gls{lrp}-$\varepsilon$ is supported, and there is no interface for custom rules or rule-maps.
Since there is no support for other \gls{lrp} rules, there is also no Canonization.

TorchRay~\citep{fong2019understanding} is another alternative that implements attribution methods in PyTorch.
It does not support any propagation-based approaches except for Guided Backprop and related methods.
Although other attribution methods are supported, most notably RISE \cite{petsiuk2018rise}, the project is currently unmaintained.

Finally, DeepExplain~\citep{ancona2018towards} provides another alternative in Keras.
DeepExplain only supports \gls{lrp}-$\varepsilon$ and DeepLIFT for propagation-based attribution.
While some other attribution-based approaches are available which are not supported in iNNvestigate and tests with continuous integration are implemented,
its documentation is limited and the framework is currently unmaintained.

\paragraph{Pipelining Frameworks}
\corelay mainly focuses on the implementation of pipelines related to \gls{spray},
and thus is specifically designed to provide compatible data to \virelay.
Although it was made with a focus on this specific use-case, the same workflow can be implemented using alternative pipelining frameworks, which we compare.
Since \corelay uses implementations provided by Scikit-Learn \citep{pedregosa2011scikit} for some steps (e.g. t-SNE and k-means) in its pipeline,
a logical alternative is to use Scikit-Learn's native pipelining framework.
Similar to \corelay, Scikit-Learn's pipelining framework is optimized for single machine pipelines implemented in Python.
Both frameworks come with the ability to cache intermediate results.
A down-side of using Scikit-Learn's pipelines is the increased implementation cost for \gls{spray} and the necessary implementation of the interface to \virelay .

Luigi \citep{bernhardsson2012luigi} is a more advanced pipelining framework specifically made for long-running batch jobs.
While \corelay and Scikit-Learn generally use parts of computations as tasks in a single pipeline,
Luigi sits one layer of abstraction higher, where it delegates (generally up to thousands of) tasks in multiple pipelines,
which are not necessarily only computations in Python.
It comes with a client-server model, where a central server schedules tasks executed by clients.
Furthermore, a web server is built into Luigi to visualize the dependency graph.
While smaller pipelines like \gls{spray} could be constructed and executed using Luigi,
its computational complexity, even on data as large as ImageNet \citep{russakovsky2015imagenet}, is generally low enough to allow the use of only a single machine.

For even more advanced, distributed pipelines, Apache AirFlow \citep{beauchemin2014airflow} can be used to develop, schedule and monitor complex batch-jobs.
Although similar to Luigi in functionality, AirFlow provides a large amount of interoperability and integration for distributed and high-performance computing, as well as high scalability.
While suitable for both large and small workflows, including \gls{spray}, the added code complexity may outweigh its benefits for small, single-machine workflows specifically.

\paragraph{XAI Visualization Frameworks}
\virelay solves a specific problem by visualizing image samples along an auxiliary visual representation (here attributions),
a 2d-representation in which samples can easily be compared, as well as multiple color-coded clusterings/ labelings.
Even though different applications are also possible,
\virelay is primarily designed to visualize the embedding and clustering of representations of attributions as provided by the extended \citep{anders2022finding} \gls{spray} approach.
Due to its application-specific nature, there are no true alternatives which do not involve the design of a new framework.
However, other frameworks designed to aid the user in the examination of models using \gls{xai} or attribution approaches exist.

Captum Insights \citep{kokhlikyan2020captum} is a web interface, directly included in Captum, to visualize and interact with data samples, model predictions, and feature attributions.
While it does not visualize embeddings or clusterings, it allows to show the data alongside attributions, somewhat similar, but more basic than \virelay , and the prediction probabilities for the different classes.
Captum Insights is mostly static, so its provided level of interaction for exploration is rather limited.

The interpretML \citep{nori2019interpretml} framework itself focuses on fitting \emph{glassbox} (i.e. inherently interpretable) models, while also providing a few post-hoc explanation methods.
It features a dashboard, where, along data exploration and model performance, feature importance of individual samples can be visually explored.
Although both interpretML's dashboard and \virelay visualize feature importance,
\virelay focuses more on the \emph{analysis} of the feature importance rather than the visualization of model performance and predictions.

Explainerdashboard \citep{dijk2022explainerdashboard} provides a similar set of explainability methods wrapped into a single, Scikit-Learn-compatible interface,
which directly executes a dashboard that provides a detailed overview of an analysis of the model.
In addition to feature importance, feature dependence, and feature interactions (provided through SHAP Values \citep{castro2009polynomial,strumbelj2010efficient}),
the dashboard shows statistics over the model performance, predictions for specific samples,
a sample perturbation interface to analyze the prediction under specific changes to individual samples,
as well as, specifically for Random Forests and xgboost models, a view of the individual decision trees.
Explainerdashboard provides good interaction to understand simple models trained on tabular data.
However, feature analysis of image data, as implemented in \virelay , is impossible.

\section{Dataset-wide Explainable AI}
In the following we demonstrate how results from \zennit and \corelay can be analyzed using \virelay to discover artifactual patterns in a model's prediction. For a technical description of the creation of a \virelay project, we refer to the supplement.

\paragraph{Analyzing Classifiers and Datasets}
\citet{lapuschkin2019unmasking} performed a \gls{spray} analysis on a Fisher vector classifier trained on the PASCAL VOC 2007 dataset and found spurious correlations in the dataset.
We will recreate this analysis and show how using \virelay simplifies the process of finding these defects in an intuitive fashion.
A complete guide for this analysis can be found in our documentation\footnote{\url{https://virelay.rtfd.io/en/0.4.0/user-guide/how-to-analyze-classifiers-and-datasets.html}}.

\begin{figure}[ht]
    \centering
    \includegraphics[width=0.75\textwidth]{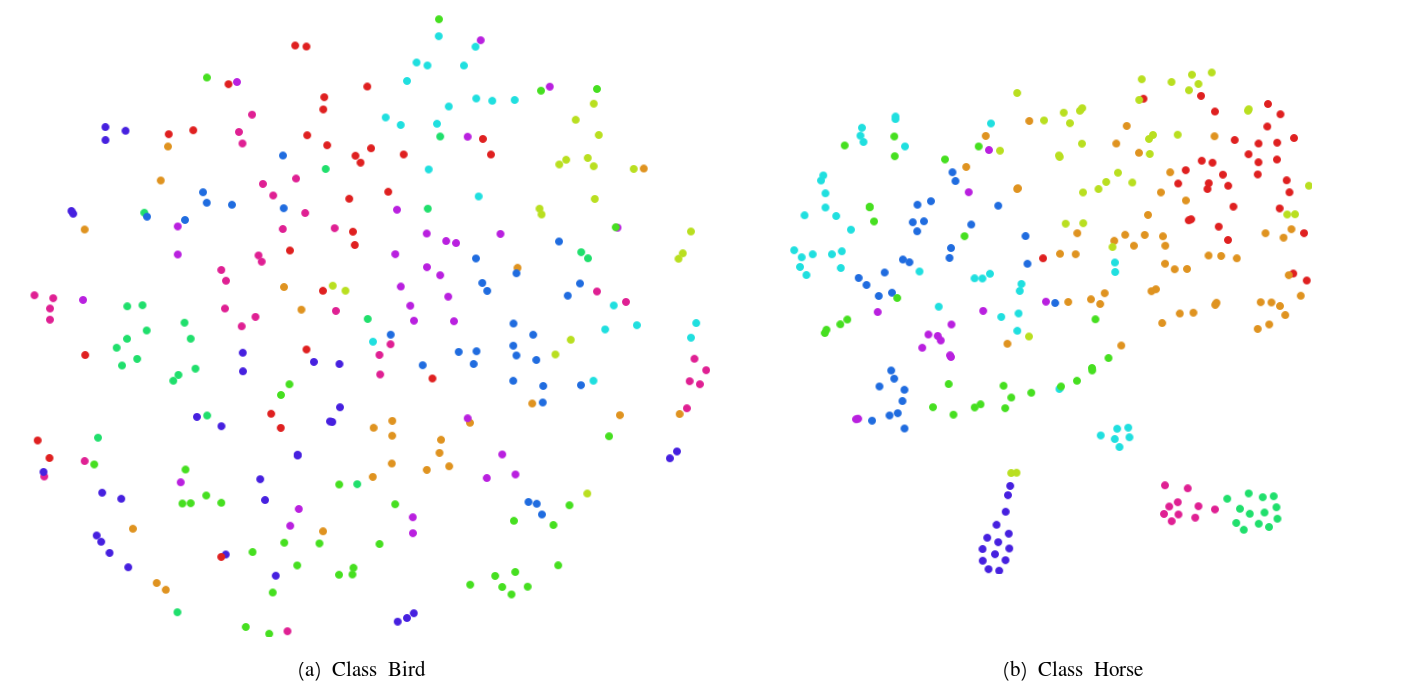}
    \caption{Comparison of \gls{tsne} embeddings of classes bird (left) and horse (right).}
    \label{fig:comparison-class-bird-vs-class-horse}
\end{figure}

We used a \gls{spray} pipeline that, based on the samples in input space, produced \gls{tsne} (as seen in Figure \ref{fig:comparison-class-bird-vs-class-horse}) and spectral embeddings of the attributions, as well as clusterings (\eg $k$-Means), similar to the implementation in Listing \ref{lst:pipeline}.
The samples were categorized by their Pascal VOC 2007 class.
When starting out with the inspection using \virelay, it is often helpful to get an overview of the different embeddings and clusterings.
Depending on the project, different input data representations (\ie in feature space of the \gls{dnn}), as well as hyper-parameters and types of embeddings and clusterings can provide more insight than others.
In this project, the \gls{tsne} embeddings based on the samples in input space turned out to be most informative, therefore, they were investigated further, to find outlier clusters.
Especially, when the outlier clusters are small, this may hint at a Clever Hans classification strategy, because the strategy was learned for only a small subset of training samples, thus suggesting that these samples may have a special feature in common that can be easily exploited by the classifier.
Indeed, when looking through the \gls{tsne} embeddings for the different classes, some classes have a very homogeneous embedding, while others have one or more outlier clusters.
For example, see Figure \ref{fig:comparison-class-bird-vs-class-horse}, which shows a comparison of the \gls{tsne} embeddings for the classes bird and horse.

\begin{figure}[ht]
    \centering
    \begin{tabular}{ m{0.15\textwidth} m{0.75\textwidth} }
    Input & \includegraphics[width=0.75\textwidth]{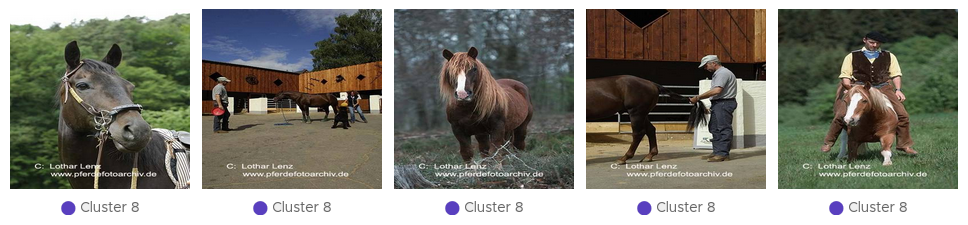} \\
    Attribution\newline Heatmap\newline Superimposed & \includegraphics[width=0.75\textwidth]{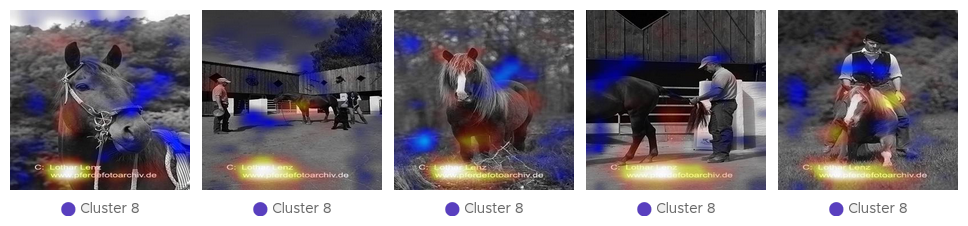}
    \end{tabular}
    \caption{
        Top: input images of the samples in an outlier cluster of the class horse;
        Bottom: same images with the attribution heatmap superimposed onto them (sample viewer display mode overlay). Positive relevance is superimposed onto a gray-scale version of each image, in ascending order, from red to yellow to white color. Negative relevance is superimposed, in ascending order, from blue to cyan.
    }
    \label{fig:input-images-of-outlier-cluster}
\end{figure}

The \gls{tsne} embedding for the class bird is very homogeneous, which can be interpreted as the classifier having learned a coherent and robust classification strategy.
On the other hand, the \gls{tsne} embedding for the class horse has multiple outlier clusters at the bottom, which means that the classifier has likely learned multiple distinct classification strategies for the class horse, each of which only apply to a distinct subset of the data domain.
The small outlier clusters suggest that those classification strategies were only learned for a few select training samples, thus warranting further manual investigation.
When visually inspecting a few training samples from the drop shaped outlier cluster at the bottom, which can be seen in Figure \ref{fig:input-images-of-outlier-cluster} (top), one common feature directly sticks out: all images have a common copyright notice at the bottom.

To validate this hypothesis, the attributions have to be surveyed.
\virelay makes it possible to directly view the attributions in the form of heatmaps or as heatmaps directly superimposed onto the input images, enabling the user to straightforwardly correlate the attributions with the underlying image features.
When the attributions are fine and detailed, they tend to get imperceptible in overlay mode and are better viewed directly.
However, when the attributions are coarse, it is harder to correlate the features in the heatmap to the corresponding image regions, therefore, the overlay mode makes it easier to see what image regions were attributed.
In this project, the attributions are rather coarse, thus making the overlay mode a better choice.
The attributions of the input images superimposed onto the input images can be seen in Figure \ref{fig:input-images-of-outlier-cluster} (bottom).

The attribution confirms the hypothesis, that the classifier bases its classification decision on the copyright notice at the bottom of the images.
In fact, it seems to almost exclusively rely on the copyright notice, as there is even some negative attribution on image regions that contain parts of the horse.
When the other outlier clusters are investigated, it turns out that they all suffer from the same artifact, although each cluster has a different copyright notice.
This indicates that the classifier has learned classification strategy for the class horse that is artifactual and does not generalize to real-world usage.
One might even conclude that these are bad training samples, as they may lead the learning algorithm astray.
Most importantly, although this particular finding was already known, this demonstrates the practicality to find such defects in classifiers and datasets using \zennit, \corelay, and \virelay.

\section{Conclusion}
In advocacy of reproducibility in machine learning~\citep{sonnenburg2007need}, we have introduced three open source software frameworks to attribute, analyze, and interactively explore a model's dataset-wide prediction strategies:
With \zennit, we hope to provide an intuitive tool within the boundaries of PyTorch to compute attributions in a customizable and intuitive fashion, and to make the multitude of rules in \gls{lrp} and other rule-based attribution methods more accessible.
We especially hope that any kind of model can now be analyzed by extending attribution approaches easily based on the intuitive structure of \zennit.
By introducing \corelay, we hope to provide a simple way to analyze attributions dataset-wide in swiftly built pipelines, and thus explore the unused potential of insight into prediction models.
Using \virelay, we hope to make the exploration of analysis results as effortless as possible by providing an interactive combined viewer of source data, attributions, visual embeddings, clusterings, and others.
\zennit, \corelay, and \virelay in combination have already been successfully used in the analysis of ImageNet~\citep{russakovsky2015imagenet} on millions of images to find artifactual Clever Hans behavior~\citep{anders2022finding}, thus demonstrating effectiveness and scalability.
With the frameworks' introduction, we hope to aid the community in the research and application of methods of \gls{xai} and beyond, to gain deeper insights into the prediction strategies of \glspl{dnn}.

\section*{Acknowledgements}
CJA, WS, and KRM were supported by the German Ministry for Education and Research (BMBF)
under grants 01IS14013A-E, 01GQ1115, 01GQ0850, 01IS18056A, 01IS18025A and 01IS18037A.
DN, WS, and SL received funding from the European Union's Horizon 2020 research and innovation programme under grant iToBoS (grant No. 965221), and
the state of Berlin within the innovation support program ProFIT (IBB) as grant BerDiBa (grant no. 10174498).
WS was further supported by the German Research Foundation (ref.\ DFG KI-FOR 5363).
KRM was also supported by the
Information \& Communications Technology Planning \& Evaluation (IITP) grant funded by the Korea government (grant no. 2017-0-001779),
as well as by the Research Training Group ``Differential Equation- and Data-driven Models in Life Sciences and Fluid Dynamics (DAEDALUS)'' (GRK 2433) and Grant Math+, EXC 2046/1, Project ID 390685689 both funded by the German Research Foundation (DFG).

\bibliographystyle{IEEEtranN}
\bibliography{main}

\begin{thebibliography}{76}
\providecommand{\natexlab}[1]{#1}
\providecommand{\url}[1]{#1}
\csname url@samestyle\endcsname
\providecommand{\newblock}{\relax}
\providecommand{\bibinfo}[2]{#2}
\providecommand{\BIBentrySTDinterwordspacing}{\spaceskip=0pt\relax}
\providecommand{\BIBentryALTinterwordstretchfactor}{4}
\providecommand{\BIBentryALTinterwordspacing}{\spaceskip=\fontdimen2\font plus
\BIBentryALTinterwordstretchfactor\fontdimen3\font minus
  \fontdimen4\font\relax}
\providecommand{\BIBforeignlanguage}[2]{{%
\expandafter\ifx\csname l@#1\endcsname\relax
\typeout{** WARNING: IEEEtranN.bst: No hyphenation pattern has been}%
\typeout{** loaded for the language `#1'. Using the pattern for}%
\typeout{** the default language instead.}%
\else
\language=\csname l@#1\endcsname
\fi
#2}}
\providecommand{\BIBdecl}{\relax}
\BIBdecl

\bibitem[Senior et~al.(2020)Senior, Evans, Jumper, Kirkpatrick, Sifre, Green,
  Qin, Z{\'{\i}}dek, Nelson, Bridgland, Penedones, Petersen, Simonyan, Crossan,
  Kohli, Jones, Silver, Kavukcuoglu, and Hassabis]{senior2020improved}
A.~W. Senior, R.~Evans, J.~Jumper, J.~Kirkpatrick, L.~Sifre, T.~Green, C.~Qin,
  A.~Z{\'{\i}}dek, A.~W.~R. Nelson, A.~Bridgland, H.~Penedones, S.~Petersen,
  K.~Simonyan, S.~Crossan, P.~Kohli, D.~T. Jones, D.~Silver, K.~Kavukcuoglu,
  and D.~Hassabis, ``Improved protein structure prediction using potentials
  from deep learning,'' \emph{Nature}, vol. 577, no. 7792, pp. 706--710, 2020.

\bibitem[Unke et~al.(2021)Unke, Chmiela, Gastegger, Sch{\"u}tt, Sauceda, and
  M{\"u}ller]{unke2021spookynet}
O.~T. Unke, S.~Chmiela, M.~Gastegger, K.~T. Sch{\"u}tt, H.~E. Sauceda, and
  K.-R. M{\"u}ller, ``Spookynet: Learning force fields with electronic degrees
  of freedom and nonlocal effects,'' \emph{Nature communications}, vol.~12,
  no.~1, p. 7273, 2021.

\bibitem[Dosovitskiy et~al.(2021)Dosovitskiy, Beyer, Kolesnikov, Weissenborn,
  Zhai, Unterthiner, Dehghani, Minderer, Heigold, Gelly, Uszkoreit, and
  Houlsby]{dosovitskiy2021image}
A.~Dosovitskiy, L.~Beyer, A.~Kolesnikov, D.~Weissenborn, X.~Zhai,
  T.~Unterthiner, M.~Dehghani, M.~Minderer, G.~Heigold, S.~Gelly, J.~Uszkoreit,
  and N.~Houlsby, ``An image is worth 16x16 words: Transformers for image
  recognition at scale,'' in \emph{9th International Conference on Learning
  Representations, {ICLR} 2021, Virtual Event, Austria, May 3-7, 2021}.\hskip
  1em plus 0.5em minus 0.4em\relax OpenReview.net, 2021.

\bibitem[Guidotti et~al.(2019)Guidotti, Monreale, Ruggieri, Turini, Giannotti,
  and Pedreschi]{guidotti2019survey}
R.~Guidotti, A.~Monreale, S.~Ruggieri, F.~Turini, F.~Giannotti, and
  D.~Pedreschi, ``A survey of methods for explaining black box models,''
  \emph{{ACM} Computing Surveys}, vol.~51, no.~5, pp. 93:1--93:42, 2019.

\bibitem[Samek et~al.(2019)Samek, Montavon, Vedaldi, Hansen, and
  M{\"{u}}ller]{samek2019explainable}
W.~Samek, G.~Montavon, A.~Vedaldi, L.~K. Hansen, and K.~M{\"{u}}ller, Eds.,
  \emph{Explainable {AI:} Interpreting, Explaining and Visualizing Deep
  Learning}, ser. Lecture Notes in Computer Science.\hskip 1em plus 0.5em minus
  0.4em\relax Springer, 2019, vol. 11700.

\bibitem[Arrieta et~al.(2020)Arrieta, Rodr{\'{\i}}guez, Ser, Bennetot, Tabik,
  Barbado, Garc{\'{\i}}a, Gil{-}Lopez, Molina, Benjamins, Chatila, and
  Herrera]{arrieta2020explainable}
A.~B. Arrieta, N.~D. Rodr{\'{\i}}guez, J.~D. Ser, A.~Bennetot, S.~Tabik,
  A.~Barbado, S.~Garc{\'{\i}}a, S.~Gil{-}Lopez, D.~Molina, R.~Benjamins,
  R.~Chatila, and F.~Herrera, ``Explainable artificial intelligence {(XAI):}
  concepts, taxonomies, opportunities and challenges toward responsible {AI},''
  \emph{Inf. Fusion}, vol.~58, pp. 82--115, 2020.

\bibitem[Samek et~al.(2021)Samek, Montavon, Lapuschkin, Anders, and
  M{\"{u}}ller]{samek2021toward}
W.~Samek, G.~Montavon, S.~Lapuschkin, C.~J. Anders, and K.-R. M{\"{u}}ller,
  ``Explaining deep neural networks and beyond: {A} review of methods and
  applications,'' \emph{Proceedings of the {IEEE}}, vol. 109, no.~3, pp.
  247--278, 2021.

\bibitem[Holzinger et~al.(2022)Holzinger, Saranti, Molnar, Biece, and
  Samek]{holzinger2022explainable}
A.~Holzinger, A.~Saranti, C.~Molnar, P.~Biece, and W.~Samek, ``Explainable ai
  methods - a brief overview,'' in \emph{xxAI - Beyond Explainable AI}, ser.
  Lecture Notes in Artificial Intelligence, A.~Holzinger, R.~Goebel, R.~Fong,
  T.~Moon, K.-R. M{\"u}ller, and W.~Samek, Eds., 2022, vol. 13200, pp. 13--38.

\bibitem[Bach et~al.(2015)Bach, Binder, Montavon, Klauschen, M{\"u}ller, and
  Samek]{bach2015pixel}
S.~Bach, A.~Binder, G.~Montavon, F.~Klauschen, K.-R. M{\"u}ller, and W.~Samek,
  ``On pixel-wise explanations for non-linear classifier decisions by
  layer-wise relevance propagation,'' \emph{PLoS ONE}, vol.~10, no.~7, p.
  e0130140, 2015.

\bibitem[Montavon et~al.(2019)Montavon, Binder, Lapuschkin, Samek, and
  M{\"u}ller]{montavon2019layer}
G.~Montavon, A.~Binder, S.~Lapuschkin, W.~Samek, and K.-R. M{\"u}ller,
  ``Layer-wise relevance propagation: an overview,'' in \emph{Explainable AI:
  Interpreting, Explaining and Visualizing Deep Learning}.\hskip 1em plus 0.5em
  minus 0.4em\relax Springer LNCS 11700, 2019, pp. 193--209.

\bibitem[Montavon et~al.(2017)Montavon, Lapuschkin, Binder, Samek, and
  M{\"u}ller]{montavon2017explaining}
G.~Montavon, S.~Lapuschkin, A.~Binder, W.~Samek, and K.-R. M{\"u}ller,
  ``Explaining nonlinear classification decisions with deep taylor
  decomposition,'' \emph{Pattern Recognition}, vol.~65, pp. 211--222, 2017.

\bibitem[Kohlbrenner et~al.(2020)Kohlbrenner, Bauer, Nakajima, Binder, Samek,
  and Lapuschkin]{kohlbrenner2020towards}
M.~Kohlbrenner, A.~Bauer, S.~Nakajima, A.~Binder, W.~Samek, and S.~Lapuschkin,
  ``Towards best practice in explaining neural network decisions with lrp,'' in
  \emph{Proceedings of the IEEE International Joint Conference on Neural
  Networks (IJCNN)}, 2020, pp. 1--7.

\bibitem[Arras et~al.(2019)Arras, Arjona{-}Medina, Widrich, Montavon,
  Gillhofer, M{\"{u}}ller, Hochreiter, and Samek]{arras2019explaining}
L.~Arras, J.~A. Arjona{-}Medina, M.~Widrich, G.~Montavon, M.~Gillhofer, K.-R.
  M{\"{u}}ller, S.~Hochreiter, and W.~Samek, ``Explaining and interpreting
  lstms,'' in \emph{Explainable {AI:} Interpreting, Explaining and Visualizing
  Deep Learning}, ser. Lecture Notes in Computer Science.\hskip 1em plus 0.5em
  minus 0.4em\relax Springer, 2019, vol. 11700, pp. 211--238.

\bibitem[Samek et~al.(2017)Samek, Binder, Montavon, Lapuschkin, and
  M{\"u}ller]{samek2017evaluating}
W.~Samek, A.~Binder, G.~Montavon, S.~Lapuschkin, and K.-R. M{\"u}ller,
  ``Evaluating the visualization of what a deep neural network has learned,''
  \emph{IEEE Transactions on Neural Networks and Learning Systems}, vol.~28,
  no.~11, pp. 2660--2673, 2017.

\bibitem[P{\"{o}}rner et~al.(2018)P{\"{o}}rner, Sch{\"{u}}tze, and
  Roth]{poerner2018evaluating}
N.~P{\"{o}}rner, H.~Sch{\"{u}}tze, and B.~Roth, ``Evaluating neural network
  explanation methods using hybrid documents and morphosyntactic agreement,''
  in \emph{Proceedings of the Association for Computational Linguistics,
  (ACL)}.\hskip 1em plus 0.5em minus 0.4em\relax Association for Computational
  Linguistics, 2018, pp. 340--350.

\bibitem[Arras et~al.(2022)Arras, Osman, and Samek]{arras2020towards}
L.~Arras, A.~Osman, and W.~Samek, ``Clevr-xai: A benchmark dataset for the
  ground truth evaluation of neural network explanations,'' \emph{Information
  Fusion}, vol.~81, pp. 14--40, 2022.

\bibitem[Kokhlikyan et~al.(2020)Kokhlikyan, Miglani, Martin, Wang, Alsallakh,
  Reynolds, Melnikov, Kliushkina, Araya, Yan, and
  Reblitz{-}Richardson]{kokhlikyan2020captum}
N.~Kokhlikyan, V.~Miglani, M.~Martin, E.~Wang, B.~Alsallakh, J.~Reynolds,
  A.~Melnikov, N.~Kliushkina, C.~Araya, S.~Yan, and O.~Reblitz{-}Richardson,
  ``Captum: {A} unified and generic model interpretability library for
  pytorch,'' \emph{CoRR}, vol. abs/2009.07896, 2020.

\bibitem[Lapuschkin et~al.(2016{\natexlab{a}})Lapuschkin, Binder, Montavon,
  M\"uller, and Samek]{lapuschkin2016analyzing}
S.~Lapuschkin, A.~Binder, G.~Montavon, K.-R. M\"uller, and W.~Samek,
  ``Analyzing classifiers: Fisher vectors and deep neural networks,'' in
  \emph{Proceedings of the IEEE Conference on Computer Vision and Pattern
  Recognition (CVPR)}, 2016, pp. 2912--2920.

\bibitem[Aeles et~al.(2021)Aeles, Horst, Lapuschkin, Lacourpaille, and
  Hug]{aeles2021revealing}
J.~Aeles, F.~Horst, S.~Lapuschkin, L.~Lacourpaille, and F.~Hug, ``Revealing the
  unique features of each individual's muscle activation signatures,''
  \emph{Journal of the Royal Society Interface}, vol.~18, no. 174, p. 20200770,
  2021.

\bibitem[Pfungst(1911)]{pfungst1911clever}
O.~Pfungst, \emph{Clever Hans: (the horse of Mr. Von Osten.) a contribution to
  experimental animal and human psychology}.\hskip 1em plus 0.5em minus
  0.4em\relax Holt, Rinehart and Winston, 1911.

\bibitem[Lapuschkin et~al.(2019)Lapuschkin, W{\"a}ldchen, Binder, Montavon,
  Samek, and M{\"u}ller]{lapuschkin2019unmasking}
S.~Lapuschkin, S.~W{\"a}ldchen, A.~Binder, G.~Montavon, W.~Samek, and K.-R.
  M{\"u}ller, ``Unmasking clever hans predictors and assessing what machines
  really learn,'' \emph{Nature Communications}, vol.~10, p. 1096, 2019.

\bibitem[Anders et~al.(2022)Anders, Weber, Neumann, Samek, M{\"{u}}ller, and
  Lapuschkin]{anders2022finding}
C.~J. Anders, L.~Weber, D.~Neumann, W.~Samek, K.-R. M{\"{u}}ller, and
  S.~Lapuschkin, ``Finding and removing clever hans: Using explanation methods
  to debug and improve deep models,'' \emph{Information Fusion}, vol.~77, pp.
  261--295, 2022.

\bibitem[Lapuschkin et~al.(2016{\natexlab{b}})Lapuschkin, Binder, Montavon,
  M{\"{u}}ller, and Samek]{lapuschkin2016toolbox}
S.~Lapuschkin, A.~Binder, G.~Montavon, K.-R. M{\"{u}}ller, and W.~Samek, ``The
  {LRP} toolbox for artificial neural networks,'' \emph{Journal of Machine
  Learning Research}, vol.~17, pp. 114:1--114:5, 2016.

\bibitem[Jia et~al.(2014)Jia, Shelhamer, Donahue, Karayev, Long, Girshick,
  Guadarrama, and Darrell]{jia2014caffe}
Y.~Jia, E.~Shelhamer, J.~Donahue, S.~Karayev, J.~Long, R.~Girshick,
  S.~Guadarrama, and T.~Darrell, ``Caffe: Convolutional architecture for fast
  feature embedding,'' in \emph{Proceedings of the 22nd ACM international
  conference on Multimedia}, 2014, pp. 675--678.

\bibitem[Harris et~al.(2020)Harris, Millman, van~der Walt, Gommers, Virtanen,
  Cournapeau, Wieser, Taylor, Berg, Smith, Kern, Picus, Hoyer, van Kerkwijk,
  Brett, Haldane, Fernández~del Río, Wiebe, Peterson, Gérard-Marchant,
  Sheppard, Reddy, Weckesser, Abbasi, Gohlke, and Oliphant]{harris2020array}
C.~R. Harris, K.~J. Millman, S.~J. van~der Walt, R.~Gommers, P.~Virtanen,
  D.~Cournapeau, E.~Wieser, J.~Taylor, S.~Berg, N.~J. Smith, R.~Kern, M.~Picus,
  S.~Hoyer, M.~H. van Kerkwijk, M.~Brett, A.~Haldane, J.~Fernández~del Río,
  M.~Wiebe, P.~Peterson, P.~Gérard-Marchant, K.~Sheppard, T.~Reddy,
  W.~Weckesser, H.~Abbasi, C.~Gohlke, and T.~E. Oliphant, ``Array programming
  with {NumPy},'' \emph{Nature}, vol. 585, p. 357–362, 2020.

\bibitem[Okuta et~al.(2017)Okuta, Unno, Nishino, Hido, and
  Loomis]{okuta2017cupy}
R.~Okuta, Y.~Unno, D.~Nishino, S.~Hido, and C.~Loomis, ``Cupy: A
  numpy-compatible library for nvidia gpu calculations,'' in \emph{Proceedings
  of Workshop on Machine Learning Systems (LearningSys) in The Thirty-first
  Annual Conference on Neural Information Processing Systems (NIPS)}, 2017.

\bibitem[Alber et~al.(2019)Alber, Lapuschkin, Seegerer, H{\"{a}}gele,
  Sch{\"{u}}tt, Montavon, Samek, M{\"{u}}ller, D{\"{a}}hne, and
  Kindermans]{alber2019innvestigate}
M.~Alber, S.~Lapuschkin, P.~Seegerer, M.~H{\"{a}}gele, K.~T. Sch{\"{u}}tt,
  G.~Montavon, W.~Samek, K.-R. M{\"{u}}ller, S.~D{\"{a}}hne, and P.-J.
  Kindermans, ``{iNNvestigate Neural Networks!}'' \emph{Journal of Machine
  Learning Research}, vol.~20, pp. 93:1--93:8, 2019.

\bibitem[Abadi et~al.(2016)Abadi, Barham, Chen, Chen, Davis, Dean, Devin,
  Ghemawat, Irving, Isard, et~al.]{abadi2016tensorflow}
M.~Abadi, P.~Barham, J.~Chen, Z.~Chen, A.~Davis, J.~Dean, M.~Devin,
  S.~Ghemawat, G.~Irving, M.~Isard \emph{et~al.}, ``Tensorflow: A system for
  large-scale machine learning,'' in \emph{$\{$USENIX$\}$ Symposium on
  Operating Systems Design and Implementation ($\{$OSDI$\}$ 16)}, 2016, pp.
  265--283.

\bibitem[Chollet et~al.(2015)]{chollet2015keras}
\BIBentryALTinterwordspacing
F.~Chollet \emph{et~al.} (2015) Keras. [Online]. Available:
  \url{https://github.com/fchollet/keras}
\BIBentrySTDinterwordspacing

\bibitem[Fong et~al.(2019)Fong, Patrick, and Vedaldi]{fong2019understanding}
R.~Fong, M.~Patrick, and A.~Vedaldi, ``Understanding deep networks via extremal
  perturbations and smooth masks,'' in \emph{2019 {IEEE/CVF} International
  Conference on Computer Vision, {ICCV} 2019, Seoul, Korea (South), October 27
  - November 2, 2019}.\hskip 1em plus 0.5em minus 0.4em\relax {IEEE}, 2019, pp.
  2950--2958.

\bibitem[Agarwal et~al.(2022)Agarwal, Krishna, Saxena, Pawelczyk, Johnson,
  Puri, Zitnik, and Lakkaraju]{agarwal2022openxai}
C.~Agarwal, S.~Krishna, E.~Saxena, M.~Pawelczyk, N.~Johnson, I.~Puri,
  M.~Zitnik, and H.~Lakkaraju, ``Open{XAI}: Towards a transparent evaluation of
  model explanations,'' in \emph{Thirty-sixth Conference on Neural Information
  Processing Systems Datasets and Benchmarks Track}, 2022.

\bibitem[Hedstr{\"{o}}m et~al.(2023)Hedstr{\"{o}}m, Weber, Krakowczyk, Bareeva,
  Motzkus, Samek, Lapuschkin, and H{\"{o}}hne]{hedstrom2023quantus}
A.~Hedstr{\"{o}}m, L.~Weber, D.~Krakowczyk, D.~Bareeva, F.~Motzkus, W.~Samek,
  S.~Lapuschkin, and M.~M.~M. H{\"{o}}hne, ``Quantus: An explainable ai toolkit
  for responsible evaluation of neural network explanations and beyond,''
  \emph{Journal of Machine Learning Research}, vol.~24, no.~34, pp. 1--11,
  2023.

\bibitem[Achtibat et~al.(2022)Achtibat, Dreyer, Eisenbraun, Bosse, Wiegand,
  Samek, and Lapuschkin]{achtibat2022from}
R.~Achtibat, M.~Dreyer, I.~Eisenbraun, S.~Bosse, T.~Wiegand, W.~Samek, and
  S.~Lapuschkin, ``From "where" to "what": Towards human-understandable
  explanations through concept relevance propagation,'' \emph{CoRR}, vol.
  abs/2206.03208, 2022.

\bibitem[Paszke et~al.(2019)Paszke, Gross, Massa, Lerer, Bradbury, Chanan,
  Killeen, Lin, Gimelshein, Antiga, Desmaison, K{\"{o}}pf, Yang, DeVito,
  Raison, Tejani, Chilamkurthy, Steiner, Fang, Bai, and
  Chintala]{paszke2019pytorch}
A.~Paszke, S.~Gross, F.~Massa, A.~Lerer, J.~Bradbury, G.~Chanan, T.~Killeen,
  Z.~Lin, N.~Gimelshein, L.~Antiga, A.~Desmaison, A.~K{\"{o}}pf, E.~Yang,
  Z.~DeVito, M.~Raison, A.~Tejani, S.~Chilamkurthy, B.~Steiner, L.~Fang,
  J.~Bai, and S.~Chintala, ``Pytorch: An imperative style, high-performance
  deep learning library,'' in \emph{Advances in Neural Information Processing
  Systems (NeurIPS)}, 2019, pp. 8024--8035.

\bibitem[Smilkov et~al.(2017)Smilkov, Thorat, Kim, Vi{\'{e}}gas, and
  Wattenberg]{smilkov2017smoothgrad}
D.~Smilkov, N.~Thorat, B.~Kim, F.~B. Vi{\'{e}}gas, and M.~Wattenberg,
  ``Smoothgrad: removing noise by adding noise,'' \emph{CoRR}, vol.
  abs/1706.03825, 2017.

\bibitem[Sundararajan et~al.(2017)Sundararajan, Taly, and
  Yan]{sundararajan2017axiomatic}
M.~Sundararajan, A.~Taly, and Q.~Yan, ``Axiomatic attribution for deep
  networks,'' in \emph{Proceedings of the 34th International Conference on
  Machine Learning, {ICML} 2017, Sydney, NSW, Australia, 6-11 August 2017},
  ser. Proceedings of Machine Learning Research, D.~Precup and Y.~W. Teh, Eds.,
  vol.~70.\hskip 1em plus 0.5em minus 0.4em\relax {PMLR}, 2017, pp. 3319--3328.

\bibitem[Springenberg et~al.(2015)Springenberg, Dosovitskiy, Brox, and
  Riedmiller]{springenberg2015striving}
J.~T. Springenberg, A.~Dosovitskiy, T.~Brox, and M.~A. Riedmiller, ``Striving
  for simplicity: The all convolutional net,'' in \emph{Proceedings of the
  International Conference of Learning Representations (ICLR)}, 2015.

\bibitem[Zhang et~al.(2018)Zhang, Bargal, Lin, Brandt, Shen, and
  Sclaroff]{zhang2018neural}
J.~Zhang, S.~A. Bargal, Z.~Lin, J.~Brandt, X.~Shen, and S.~Sclaroff, ``Top-down
  neural attention by excitation backprop,'' \emph{International Journal of
  Computer Vision}, vol. 126, no.~10, pp. 1084--1102, 2018.

\bibitem[Yeom et~al.(2021)Yeom, Seegerer, Lapuschkin, Binder, Wiedemann,
  M{\"{u}}ller, and Samek]{yeom2021pruning}
S.~Yeom, P.~Seegerer, S.~Lapuschkin, A.~Binder, S.~Wiedemann, K.-R.
  M{\"{u}}ller, and W.~Samek, ``Pruning by explaining: {A} novel criterion for
  deep neural network pruning,'' \emph{Pattern Recognition}, vol. 115, p.
  107899, 2021.

\bibitem[Ruff et~al.(2021)Ruff, Kauffmann, Vandermeulen, Montavon, Samek,
  Kloft, Dietterich, and M{\"{u}}ller]{ruff2021unifying}
L.~Ruff, J.~R. Kauffmann, R.~A. Vandermeulen, G.~Montavon, W.~Samek, M.~Kloft,
  T.~G. Dietterich, and K.-R. M{\"{u}}ller, ``A unifying review of deep and
  shallow anomaly detection,'' \emph{Proceedings of the {IEEE}}, vol. 109,
  no.~5, pp. 756--795, 2021.

\bibitem[Motzkus et~al.(2022)Motzkus, Weber, and
  Lapuschkin]{motzkus2022measurably}
F.~Motzkus, L.~Weber, and S.~Lapuschkin, ``Measurably stronger explanation
  reliability via model canonization,'' in \emph{2022 IEEE International
  Conference on Image Processing (ICIP)}.\hskip 1em plus 0.5em minus
  0.4em\relax IEEE, 2022, pp. 516--520.

\bibitem[Pahde et~al.(2022)Pahde, Yolcu, Binder, Samek, and
  Lapuschkin]{pahde2022optimizing}
F.~Pahde, G.~{\"{U}}. Yolcu, A.~Binder, W.~Samek, and S.~Lapuschkin,
  ``Optimizing explanations by network canonization and hyperparameter
  search,'' \emph{CoRR}, vol. abs/2211.17174, 2022.

\bibitem[Ioffe and Szegedy(2015)]{ioffe2015batch}
S.~Ioffe and C.~Szegedy, ``Batch normalization: Accelerating deep network
  training by reducing internal covariate shift,'' in \emph{Proceedings of the
  International Conference on Machine Learning, (ICML)}, ser. {JMLR} Workshop
  and Conference Proceedings, vol.~37.\hskip 1em plus 0.5em minus 0.4em\relax
  JMLR.org, 2015, pp. 448--456.

\bibitem[Hui and Binder(2019)]{hui2019batchnorm}
L.~Y.~W. Hui and A.~Binder, ``Batchnorm decomposition for deep neural network
  interpretation,'' in \emph{Proceedings of the International Work-Conference
  on Artificial Neural Networks (IWANN)}, ser. Lecture Notes in Computer
  Science, vol. 11507.\hskip 1em plus 0.5em minus 0.4em\relax Springer, 2019,
  pp. 280--291.

\bibitem[Alber(2019)]{alber2019efficient}
M.~Alber, ``Efficient learning machines: From kernel methods to deep
  learning,'' Ph.D. dissertation, Technical University of Berlin, Germany,
  2019.

\bibitem[Guillemot et~al.(2020)Guillemot, Heusele, Korichi, Schnebert, and
  Chen]{guillemot2020breaking}
M.~Guillemot, C.~Heusele, R.~Korichi, S.~Schnebert, and L.~Chen, ``Breaking
  batch normalization for better explainability of deep neural networks through
  layer-wise relevance propagation,'' \emph{CoRR}, vol. abs/2002.11018, 2020.

\bibitem[He et~al.(2016)He, Zhang, Ren, and Sun]{he2016deep}
K.~He, X.~Zhang, S.~Ren, and J.~Sun, ``Deep residual learning for image
  recognition,'' in \emph{Proceedings of the Conference on Computer Vision and
  Pattern Recognition (CVPR)}.\hskip 1em plus 0.5em minus 0.4em\relax {IEEE}
  Computer Society, 2016, pp. 770--778.

\bibitem[Zagoruyko and Komodakis(2016)]{zagoruyko2016wide}
S.~Zagoruyko and N.~Komodakis, ``Wide residual networks,'' in \emph{Proceedings
  of the British Machine Vision Conference (BMVC)}.\hskip 1em plus 0.5em minus
  0.4em\relax {BMVA} Press, 2016.

\bibitem[Simonyan and Zisserman(2014)]{simonyan2014very}
K.~Simonyan and A.~Zisserman, ``Very deep convolutional networks for
  large-scale image recognition,'' \emph{CoRR}, vol. abs/1409.1556, 2014.

\bibitem[Marcel and Rodriguez(2010)]{marcel2010torchvision}
S.~Marcel and Y.~Rodriguez, ``Torchvision the machine-vision package of
  torch,'' in \emph{Proceedings of the International Conference on Multimedia
  (ACM Multimedia)}.\hskip 1em plus 0.5em minus 0.4em\relax {ACM}, 2010, pp.
  1485--1488.

\bibitem[Zeiler and Fergus(2014)]{zeiler2014visualizing}
M.~D. Zeiler and R.~Fergus, ``Visualizing and understanding convolutional
  networks,'' in \emph{European conference on computer vision}.\hskip 1em plus
  0.5em minus 0.4em\relax Springer, 2014, pp. 818--833.

\bibitem[Meila and Shi(2001)]{meila2001random}
M.~Meila and J.~Shi, ``A random walks view of spectral segmentation,'' in
  \emph{Proceedings of the International Workshop on Artificial Intelligence
  and Statistics (AISTATS)}, 2001.

\bibitem[Ng et~al.(2002)Ng, Jordan, and Weiss]{ng2002spectral}
A.~Y. Ng, M.~I. Jordan, and Y.~Weiss, ``On spectral clustering: Analysis and an
  algorithm,'' in \emph{Advances in Neural Information Processing Systems},
  2002, pp. 849--856.

\bibitem[Maaten and Hinton(2008)]{maaten2008visualizing}
L.~v.~d. Maaten and G.~Hinton, ``Visualizing data using {t-SNE},''
  \emph{Journal of Machine Learning Research}, vol.~9, no. Nov, pp. 2579--2605,
  2008.

\bibitem[Grinberg(2014)]{grinberg2014flask}
M.~Grinberg, \emph{Flask Web Development - Developing Web Applications with
  Python}.\hskip 1em plus 0.5em minus 0.4em\relax O'Reilly, 2014.

\bibitem[Jain et~al.(2014)Jain, Bhansali, and Mehta]{jain2014angularjs}
N.~Jain, A.~Bhansali, and D.~Mehta, ``Angularjs: A modern mvc framework in
  javascript,'' \emph{Journal of Global Research in Computer Science}, vol.~5,
  no.~12, pp. 17--23, 2014.

\bibitem[Fortner(1998)]{fortner1998hdf}
B.~Fortner, ``Hdf: The hierarchical data format,'' \emph{Dr Dobb's J Software
  Tools Prof Program}, vol.~23, no.~5, p.~42, 1998.

\bibitem[Nori et~al.(2019)Nori, Jenkins, Koch, and
  Caruana]{nori2019interpretml}
H.~Nori, S.~Jenkins, P.~Koch, and R.~Caruana, ``Interpretml: {A} unified
  framework for machine learning interpretability,'' \emph{CoRR}, vol.
  abs/1909.09223, 2019.

\bibitem[Dijk et~al.(2019)Dijk, Bell, Gädke, Serna, Okumus,
  et~al.]{dijk2022explainerdashboard}
\BIBentryALTinterwordspacing
O.~Dijk, R.~Bell, A.~Gädke, B.~Serna, T.~Okumus \emph{et~al.},
  ``Explainerdashboard,'' 2019. [Online]. Available:
  \url{https://github.com/oegedijk/explainerdashboard}
\BIBentrySTDinterwordspacing

\bibitem[Klaise et~al.(2021)Klaise, Looveren, Vacanti, and
  Coca]{klaise2021alibi}
J.~Klaise, A.~V. Looveren, G.~Vacanti, and A.~Coca, ``Alibi explain: Algorithms
  for explaining machine learning models,'' \emph{J. Mach. Learn. Res.},
  vol.~22, pp. 181:1--181:7, 2021.

\bibitem[Shrikumar et~al.(2017)Shrikumar, Greenside, and
  Kundaje]{shrikumar2017learning}
A.~Shrikumar, P.~Greenside, and A.~Kundaje, ``Learning important features
  through propagating activation differences,'' in \emph{Proceedings of the
  34th International Conference on Machine Learning}, ser. Proceedings of
  Machine Learning Research, vol.~70.\hskip 1em plus 0.5em minus 0.4em\relax
  PMLR, 06--11 Aug 2017, pp. 3145--3153.

\bibitem[Lundberg and Lee(2017)]{lundberg2017unified}
S.~M. Lundberg and S.~Lee, ``A unified approach to interpreting model
  predictions,'' in \emph{Advances in Neural Information Processing Systems 30:
  Annual Conference on Neural Information Processing Systems 2017, December
  4-9, 2017, Long Beach, CA, {USA}}, I.~Guyon, U.~von Luxburg, S.~Bengio, H.~M.
  Wallach, R.~Fergus, S.~V.~N. Vishwanathan, and R.~Garnett, Eds., 2017, pp.
  4765--4774.

\bibitem[Dhamdhere et~al.(2019)Dhamdhere, Sundararajan, and
  Yan]{dhamdhere2019important}
K.~Dhamdhere, M.~Sundararajan, and Q.~Yan, ``How important is a neuron,'' in
  \emph{7th International Conference on Learning Representations, {ICLR} 2019,
  New Orleans, LA, USA, May 6-9, 2019}.\hskip 1em plus 0.5em minus 0.4em\relax
  OpenReview.net, 2019.

\bibitem[Shrikumar et~al.(2018)Shrikumar, Su, and
  Kundaje]{shrikumar2018computationally}
A.~Shrikumar, J.~Su, and A.~Kundaje, ``Computationally efficient measures of
  internal neuron importance,'' \emph{CoRR}, vol. abs/1807.09946, 2018.

\bibitem[Selvaraju et~al.(2020)Selvaraju, Cogswell, Das, Vedantam, Parikh, and
  Batra]{selvaraju2020grad}
R.~R. Selvaraju, M.~Cogswell, A.~Das, R.~Vedantam, D.~Parikh, and D.~Batra,
  ``Grad-cam: Visual explanations from deep networks via gradient-based
  localization,'' \emph{Int. J. Comput. Vis.}, vol. 128, no.~2, pp. 336--359,
  2020.

\bibitem[Ribeiro et~al.(2016)Ribeiro, Singh, and Guestrin]{ribeiro2016should}
M.~T. Ribeiro, S.~Singh, and C.~Guestrin, ``"why should {I} trust you?":
  Explaining the predictions of any classifier,'' in \emph{Proceedings of the
  22nd {ACM} {SIGKDD} International Conference on Knowledge Discovery and Data
  Mining, San Francisco, CA, USA, August 13-17, 2016}, B.~Krishnapuram,
  M.~Shah, A.~J. Smola, C.~C. Aggarwal, D.~Shen, and R.~Rastogi, Eds.\hskip 1em
  plus 0.5em minus 0.4em\relax {ACM}, 2016, pp. 1135--1144.

\bibitem[Castro et~al.(2009)Castro, G{\'{o}}mez, and
  Tejada]{castro2009polynomial}
J.~Castro, D.~G{\'{o}}mez, and J.~Tejada, ``Polynomial calculation of the
  shapley value based on sampling,'' \emph{Comput. Oper. Res.}, vol.~36, no.~5,
  pp. 1726--1730, 2009.

\bibitem[Strumbelj and Kononenko(2010)]{strumbelj2010efficient}
E.~Strumbelj and I.~Kononenko, ``An efficient explanation of individual
  classifications using game theory,'' \emph{J. Mach. Learn. Res.}, vol.~11,
  pp. 1--18, 2010.

\bibitem[Petsiuk et~al.(2018)Petsiuk, Das, and Saenko]{petsiuk2018rise}
V.~Petsiuk, A.~Das, and K.~Saenko, ``{RISE:} randomized input sampling for
  explanation of black-box models,'' in \emph{British Machine Vision Conference
  2018, {BMVC} 2018, Newcastle, UK, September 3-6, 2018}.\hskip 1em plus 0.5em
  minus 0.4em\relax {BMVA} Press, 2018, p. 151.

\bibitem[Kindermans et~al.(2018)Kindermans, Sch{\"{u}}tt, Alber, M{\"{u}}ller,
  Erhan, Kim, and D{\"{a}}hne]{kindermans2018learning}
P.~Kindermans, K.~T. Sch{\"{u}}tt, M.~Alber, K.-R. M{\"{u}}ller, D.~Erhan,
  B.~Kim, and S.~D{\"{a}}hne, ``Learning how to explain neural networks:
  Patternnet and patternattribution,'' in \emph{6th International Conference on
  Learning Representations, {ICLR} 2018, Vancouver, BC, Canada, April 30 - May
  3, 2018, Conference Track Proceedings}.\hskip 1em plus 0.5em minus
  0.4em\relax OpenReview.net, 2018.

\bibitem[Ancona et~al.(2018)Ancona, Ceolini, {\"{O}}ztireli, and
  Gross]{ancona2018towards}
M.~Ancona, E.~Ceolini, C.~{\"{O}}ztireli, and M.~Gross, ``Towards better
  understanding of gradient-based attribution methods for deep neural
  networks,'' in \emph{6th International Conference on Learning
  Representations, {ICLR} 2018, Vancouver, BC, Canada, April 30 - May 3, 2018,
  Conference Track Proceedings}.\hskip 1em plus 0.5em minus 0.4em\relax
  OpenReview.net, 2018.

\bibitem[Pedregosa et~al.(2011)Pedregosa, Varoquaux, Gramfort, Michel, Thirion,
  Grisel, Blondel, Prettenhofer, Weiss, Dubourg, VanderPlas, Passos,
  Cournapeau, Brucher, Perrot, and Duchesnay]{pedregosa2011scikit}
F.~Pedregosa, G.~Varoquaux, A.~Gramfort, V.~Michel, B.~Thirion, O.~Grisel,
  M.~Blondel, P.~Prettenhofer, R.~Weiss, V.~Dubourg, J.~VanderPlas, A.~Passos,
  D.~Cournapeau, M.~Brucher, M.~Perrot, and E.~Duchesnay, ``Scikit-learn:
  Machine learning in python,'' \emph{J. Mach. Learn. Res.}, vol.~12, pp.
  2825--2830, 2011.

\bibitem[Bernhardsson et~al.(2012)Bernhardsson, Freider, Rouhani, Buchfuhrer,
  Poulin, Stadther, Barbans, Kransnukhin, Crobak,
  et~al.]{bernhardsson2012luigi}
\BIBentryALTinterwordspacing
E.~Bernhardsson, E.~Freider, A.~Rouhani, D.~Buchfuhrer, G.~Poulin, D.~Stadther,
  U.~Barbans, A.~Kransnukhin, J.~Crobak \emph{et~al.}, ``{Luigi},'' 2012.
  [Online]. Available: \url{https://github.com/spotify/luigi}
\BIBentrySTDinterwordspacing

\bibitem[Russakovsky et~al.(2015)Russakovsky, Deng, Su, Krause, Satheesh, Ma,
  Huang, Karpathy, Khosla, Bernstein, Berg, and Li]{russakovsky2015imagenet}
O.~Russakovsky, J.~Deng, H.~Su, J.~Krause, S.~Satheesh, S.~Ma, Z.~Huang,
  A.~Karpathy, A.~Khosla, M.~S. Bernstein, A.~C. Berg, and F.~Li, ``Imagenet
  large scale visual recognition challenge,'' \emph{International Journal of
  Computer Vision}, vol. 115, no.~3, pp. 211--252, 2015.

\bibitem[Beauchemin et~al.(2014)Beauchemin, Naik, Potiuk, Bregu{ł}a,
  Berlin-Taylor, Cunningham, Urbaszek, et~al.]{beauchemin2014airflow}
\BIBentryALTinterwordspacing
M.~Beauchemin, K.~Naik, J.~Potiuk, K.~Bregu{ł}a, A.~Berlin-Taylor,
  J.~Cunningham, T.~Urbaszek \emph{et~al.}, ``{Apache Airflow},'' 2014.
  [Online]. Available: \url{https://github.com/apache/airflow}
\BIBentrySTDinterwordspacing

\bibitem[Sonnenburg et~al.(2007)Sonnenburg, Braun, Ong, Bengio, Bottou, Holmes,
  LeCun, M{\"{u}}ller, Pereira, Rasmussen, R{\"{a}}tsch, Sch{\"{o}}lkopf,
  Smola, Vincent, Weston, and Williamson]{sonnenburg2007need}
S.~Sonnenburg, M.~L. Braun, C.~S. Ong, S.~Bengio, L.~Bottou, G.~Holmes,
  Y.~LeCun, K.-R. M{\"{u}}ller, F.~Pereira, C.~E. Rasmussen, G.~R{\"{a}}tsch,
  B.~Sch{\"{o}}lkopf, A.~J. Smola, P.~Vincent, J.~Weston, and R.~C. Williamson,
  ``The need for open source software in machine learning,'' \emph{Journal of
  Machine Learning Research}, vol.~8, pp. 2443--2466, 2007.

\end{thebibliography}

\newpage
\appendix

\section{Appendix}

\subsection{Creating a ViRelAy Project}
A \virelay project consists of (1) a dataset, containing the training samples, (2) a label map, mapping between label indices, label names, and WordNet IDs (to display label names), (3) an attribution database, containing the attribution maps computed using \zennit, (4) an analysis database, containing \corelay meta-analysis results, and (5) a project file, containing meta-data and linking the individual files.

\paragraph{Input Data} Usually, a trained model and a dataset are available from the beginning.
\virelay supports two supported formats for datasets: an image directory with label-descriptive sub-directories containing the respective samples, or an HDF5 database, in which input images are stored either in an HDF5 dataset or a group.
HDF5 datasets are multi-dimensional arrays suitable for input images with the same resolution, which are stored as a single array of shape $samples \times channel \times height \times width$ under the key \emph{data}.
HDF5 groups are similar to files in a file system and can therefore be used in cases where the input images have varying resolutions,
stored in \incode{samples} many datasets of shape $channel \times height \times width$ with unique keys inside an HDF5 group also with key \emph{data}.
The labels are also stored in an HDF5 dataset or group depending on the storage format of the input images.
In single-label datasets, the labels are stored as an array of shape $samples$, where each entry contains the label index.
In the case of multi-label datasets, the labels are stored as an array of shape $samples \times number \, of \, labels$ using a multi-hot encoding.
When the input images are stored in an HDF5 group, then the labels are also stored in an HDF5 group called \emph{label}, with the unique input image ID as key and either the label index or the multi-hot encoded array as value.
The label map is a JSON file containing an array of labels, where each label is represented by an object that contains the label index, the optional WordNet ID, and the label name.
A complete specification and examples of a label map file can be found in our documentation\footnote{\url{https://virelay.rtfd.io/en/0.4.0/contributors-guide/project-file-format.html}}.

\paragraph{Attribution Data (from Zennit) } \zennit can be utilized to compute attributions for all samples in the dataset (see Section \ref{section:attribution-with-zennit}).
In order for \virelay to load these attributions, they also have to be stored in an HDF5 database.
The format is analogous to the HDF5 input dataset format, where the key of the dataset/group containing the attributions is instead \incode{attribution}.
In addition, the attribution database contains two more HDF5 datasets/groups: \emph{label}, containing the ground-truth labels of the respective original samples, and \emph{prediction}, containing the model's predictions of the original samples.
The labels are stored in the exact same fashion as they are stored in the HDF5 database containing the input images.
The predictions are always stored as a vector similar to the multi-label case, containing the classification scores output by the model.
Each project can only contain attributions for a single attribution method, but it can contain multiple attribution databases (e.g., an attribution database could be created per class).

\paragraph{Analysis Data (from CoRelAy)} \corelay can be used to build an analysis pipeline for the attributions, such as \gls{spray} (see Section \ref{section:building-analysis-pipelines-with-corelay}).
Analysis results for \virelay are stored in HDF5 databases.
The database may contain results of multiple \corelay analysis pipelines, each of which are stored as a group in the HDF5 file, with the name of the group as a unique identifier of the corresponding analysis.
Each analysis group may contain multiple sub-keys describing different categories of attributions for which the analysis was performed.
Categories may constitute anything that splits up the data in a helpful manner.
Usually, one category is created for each class in the dataset, but the data can be also categorized otherwise, e.g., by WordNet IDs or concepts.
The category groups contain a dataset \emph{index}, which contains the indices of the samples that are in the category, and two groups, \emph{embedding} and \emph{cluster}, which contain the embeddings and clusterings computed in the analysis pipeline respectively.
Each key in the \emph{embedding} sub-group represents a different embedding method, e.g., spectral embedding or \gls{tsne}.
Each embedding can optionally have multiple attributes: (1) \emph{eigenvalue}, which contains the eigenvalues of the eigendecomposition of the embedding, (2) \emph{embedding}, which is the name of the base embedding, if the embedding is based on another embedding, and (3) \emph{index}, which are the indices of the dimensions of the base embedding that were used.
Finally, the \emph{cluster} sub-group contains the clusterings that were used to cluster the attributions.
Each key in the \emph{cluster} sub-group represents a different clustering method with different parameters, e.g., different values of $k$ for a $k$-means clustering.
Each clustering can have additional attributes, e.g., \emph{embedding}, which is the embedding that the clustering is based on, or the parameters of the clustering algorithm.
A complete specification of the different HDF5 database formats\footnote{\url{https://virelay.rtfd.io/en/0.4.0/contributors-guide/database-specification.html}} as well as a guide on how to create a \virelay project from scratch\footnote{\url{https://virelay.rtfd.io/en/0.4.0/user-guide/how-to-create-a-project.html}} can be found in our documentation.

\paragraph{Project File (for ViRelAy)} Finally, these database files are combined in a project file based on the YAML format,
which consists of a project name, a model name, a reference to the dataset file, a reference to the label map file, a reference to the attribution files, and a reference to the analysis files.
The project and model name can be chosen arbitrarily and are only used to display them, when the project is opened in \virelay, to distinguish between multiple loaded projects.
The dataset consists of (1) an arbitrary name used for informational purposes, (2) a type to choose between image directory and HDF5 for the input data, (4) a path to the input data file, (5) the input width and (6) height to rescale the images, (7) the up-sampling and (8) down-sampling approach for rescaling, and (9) the path to the label map JSON file.
The attributions property consist of (1) an attribution method, which is the name of the approach used to compute the attributions, (2) the attribution strategy to indicate whether the true label or the predicted label was attributed, and (3) a list of source files.
Finally, the analyses property is a list of analyses that were performed on the data.
Multiple analyses can be created to compare different analysis methods.
Each analysis consists of the name of the analysis method and a list of source files.
A complete specification and an example can be found in our documentation\footnote{\url{https://virelay.rtfd.io/en/0.4.0/contributors-guide/project-file-format.html}}.

\subsection{Additional Zennit Attribution Heatmaps}

\begin{figure}[t]
    \centering
    \includegraphics[width=\textwidth]{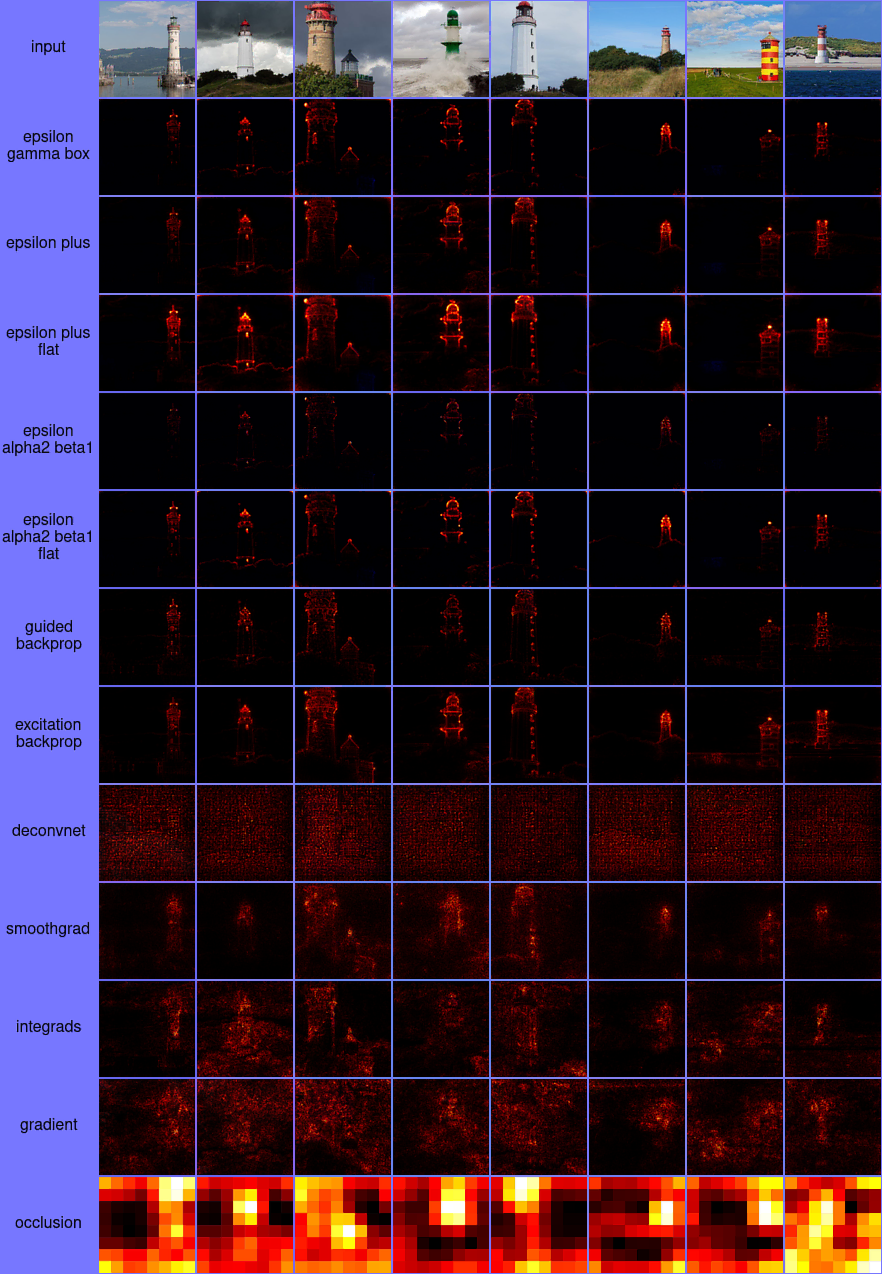}
    \caption{
        Heatmaps of attributions of lighthouses, using the pre-trained VGG-16 network with BatchNorm provided by Torchvision.
        The model correctly predicted all images as class ``lighthouse''.
        The attributions were visualized with the color map \incode{coldnhot} (negative relevance is light-/blue, irrelevant pixels are black, positive relevance is red to yellow).
    }
    \label{fig:vgg16-heatmaps}
\end{figure}

\begin{figure}[t]
    \centering
    \includegraphics[width=\textwidth]{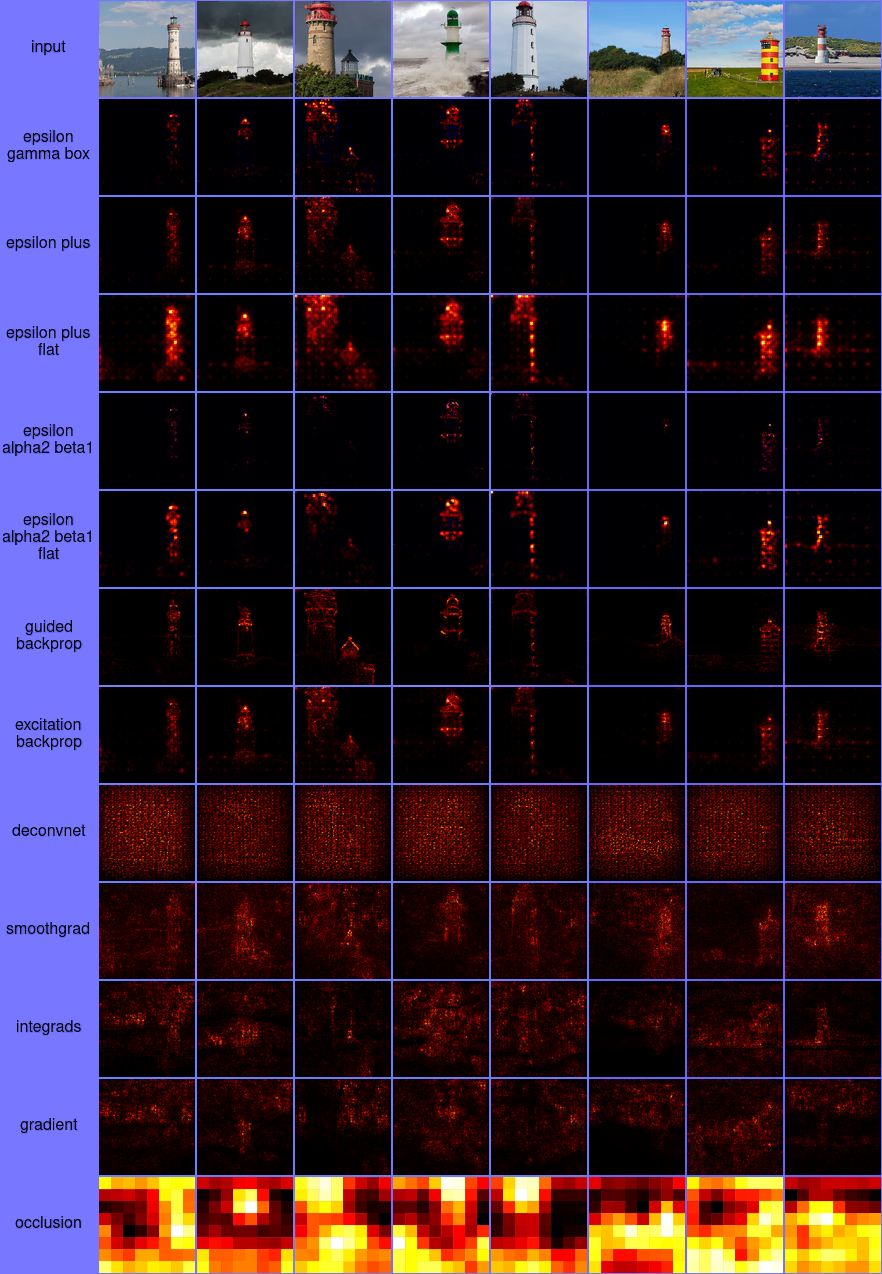}
    \caption{
        Heatmaps of attributions of lighthouses, using the pre-trained ResNet50 network provided by Torchvision.
        The model correctly predicted all images as class ``lighthouse''.
        The attributions were visualized with the color map \incode{coldnhot} (negative relevance is light-/blue, irrelevant pixels are black, positive relevance is red to yellow).
    }
    \label{fig:resnet50-heatmaps}
\end{figure}

Figure \ref{fig:vgg16-heatmaps} shows attribution heatmaps of the same Torchvision VGG16 model for various methods computed using \zennit.
Figure \ref{fig:resnet50-heatmaps} shows attribution heatmaps of Torchvision's ResNet50 model for the same methods also computed using \zennit.

\end{document}